\documentclass[11pt]{article}

\usepackage[preprint]{acl}

\usepackage{placeins}

\usepackage{times}
\usepackage{latexsym}
\usepackage{booktabs}
\usepackage{hyperref}

\usepackage[T1]{fontenc}

\usepackage[utf8]{inputenc}

\usepackage{microtype}

\usepackage{inconsolata}

\usepackage{graphicx}
\usepackage{xcolor} % Required for defining colors

\title{MOSAIC: Efficient \underline{M}ixture-\underline{o}f-Agent \underline{S}cheduling via \underline{A}daptive Aggregation and \underline{I}nference \underline{C}oncurrency}

\author{
\textbf{Saptarshi Mitra}\textsuperscript{1}\thanks{Equal contribution first authors.},
\textbf{Yifan Zhang}\textsuperscript{1}\footnotemark[1],
 \textbf{Rachid Karami\textsuperscript{1}},
  \textbf{Sreetama Sarkar\textsuperscript{2}},
  \\
 \textbf{Phyo Pyae Moe Aung\textsuperscript{1}},
 \textbf{Nazmul Takbir\textsuperscript{1}},
 \textbf{Souvik Kundu\textsuperscript{3}\thanks{Equal contribution senior authors.}},
 \textbf{Sitao Huang\textsuperscript{1}\footnotemark[2]}
\\
 \textsuperscript{1}University of California, Irvine, USA
 \\
 \textsuperscript{2}University of Southern California, Los Angeles, USA
 \textsuperscript{3}Intel, USA
\\
 \small{
   \textbf{Correspondence:} \href{mailto:saptarshi.mitra@uci.edu}{saptarshi.mitra@uci.edu}
 }
}

\usepackage{cite}
\usepackage{amsmath,amssymb,amsfonts}
\usepackage{graphicx}
\usepackage[dvipsnames]{xcolor}
\usepackage[T1]{fontenc}
\usepackage{textcomp}
\usepackage[all]{nowidow}
\usepackage[keeplastbox]{flushend}
\usepackage{fancyhdr}

\usepackage{color}
\usepackage{microtype}
\usepackage{xspace}
\usepackage{graphicx}
\usepackage{subfigure}
\usepackage{booktabs} % for professional tables

\usepackage{multirow}
\usepackage{comment}

\usepackage{hyperref}
\usepackage{bm}

\usepackage{todonotes}

\usepackage{pifont}

\usepackage{amsmath,amsfonts}
\usepackage{graphicx}
\usepackage{textcomp}
\usepackage{xcolor}
\usepackage{multirow}
\RequirePackage[nameinlink]{cleveref} % 'nameinlink' to mimic \autoref's behavior
\crefname{chapter}{Chapter}{Chapters}
\crefname{section}{Section}{Sections}
\crefname{subsection}{Section}{Sections}
\crefname{equation}{Equation}{Equations}
\crefname{definition}{Definition}{Definitions}
\crefname{assumption}{Assumption}{Assumptions}
\crefname{theorem}{Theorem}{Theorems}
\crefname{figure}{Figure}{Figures}
\crefname{table}{Table}{Tables}
\crefname{algorithm}{Algorithm}{Algorithms}
\let\autoref\cref % set \autoref as an alias for \cref

\usepackage{multirow} % Required for multirow
\usepackage{multicol} %Required for multicolumn
\usepackage{makecell} % for multi-line cell in the last two rows

\usepackage{booktabs}   % For professional table rules (\toprule, \midrule, \bottomrule)
\usepackage{tabularx}   % For tables with adjustable column widths
\usepackage{hyperref}   % For clickable URLs (\url or \href)
\usepackage{graphicx}   % Required for \resizebox if needed, or general ACM template usage
\usepackage{longtable}
\usepackage{adjustbox}
\usepackage{longtable}

\usepackage{algorithm}
\usepackage{algpseudocode}

\usepackage{enumitem}

\ifx\forsubmission\undefined
\newcommand{\TODO}[1]{\textcolor{red}{TODO: #1}}
\newcommand{\SK}[1]{\textcolor{red}{SK: #1}}

\newcommand{\HK}[1]{\textcolor{blue}{HK: #1}}
\newcommand{\RK}[1]{\textcolor{brown}{RK: #1}}

\newcommand{\fixme}[1]{{\color{red} {#1}}}

\else
\newcommand{\TODO}[1]{\textcolor{red}{}}
\newcommand{\SK}[1]{\textcolor{red}{}}
\newcommand{\FK}[1]{\textcolor{blue}{}}
\newcommand{\HK}[1]{\textcolor{blue}{}}
\newcommand{\RK}[1]{\textcolor{green}{}}

\newcommand{\fixme}[1]{{\color{red} {}}} 

\fi

\newcommand{\squishlist}{
 \begin{list}{$\bullet$}
  { \setlength{\itemsep}{0pt}
     \setlength{\parsep}{3pt}
     \setlength{\topsep}{3pt}
     \setlength{\partopsep}{0pt}
     \setlength{\leftmargin}{1.5em}
     \setlength{\labelwidth}{1em}
     \setlength{\labelsep}{0.5em} } }

\newcommand{\squishlisttwo}{
 \begin{list}{$\bullet$}
  { \setlength{\itemsep}{0pt}
     \setlength{\parsep}{0pt}
    \setlength{\topsep}{0pt}
    \setlength{\partopsep}{0pt}
    \setlength{\leftmargin}{2em}
    \setlength{\labelwidth}{1.5em}
    \setlength{\labelsep}{0.5em} } }

\newcommand{\squishend}{
  \end{list}  }

\newcommand{\betterparagraph}[1]{\noindent \textbf{#1. }}

\usepackage{graphicx}
\graphicspath{{./Figures/}}

\begin{document}
\maketitle

%% EDIT YOUR PAPER'S CONTENTS BELOW
\begin{abstract}
% <dummy abstract, must be re-written>
Mixture-of-Agents (MoA) systems improve reasoning accuracy by routing each query to multiple expert LLMs and aggregating their outputs.
Efficiently executing this workload on limited GPU resources has bottlenecks.
Skill-based routing creates skewed expert demand, and combining instruction-tuned LLMs with long-reasoning models results in extreme variability in generation lengths.
Consequently, traditional scheduling strategies suffer from significant GPU idling and throughput collapse due to load imbalances.
%
% We present MOSAIC, a
% % an offline batch 
% scheduling framework to accelerate MoA workloads.
%
We present MOSAIC, a scheduling framework for accelerating 
% offline\SM{mention offline but not highlight it}, 
batch-oriented MoA workloads.
%
% First, we formulate an Integer Linear Program (ILP) based scheduler that jointly optimizes expert placement and per-worker prompt assignment from offline-profiled costs, replicating reasoning experts across workers while pinning lightweight ones.
%
% \sout{First, an Integer Linear Program (ILP) jointly optimizes expert placement, per-worker prompt assignment, and selective replication of reasoning experts from offline-profiled costs.}
% \SM{Some alternative to 'placement level'}
MOSAIC uses two-stage workload-aware scheduling: an expert placement stage, where an integer linear program (ILP) decides expert placement, followed by a query-sequencing stage that schedules requests based on predicted output lengths.
% \YZ{MOSAIC uses two-stage workload-aware scheduling: at the expert placement stage, the algorithm decides expert placement, selective replication of experts and per-GPU prompt counts. At the question level, MOSAIC orders sessions and prompts by output lengths predicted from each expert's own hidden state.}
%
% \SM{how to connect the following adaptive aggregation}
Additionally, MOSAIC uses confidence-aware adaptive aggregation, leveraging inter-expert agreement to bypass the heavy aggregator LLM for consensus queries.
On a 4-GPU system, MOSAIC achieves $2.7$--$3.4\times$ end-to-end speedups over round-robin scheduling and reduces the mean time-to-answer by $3.9$--$7.8\times$, while matching accuracy within $\pm 1$pp.

\end{abstract}
\section{Introduction}
\label{sec:intro}

%Recent advancements in large language models (LLMs)~\citep{Touvron2023LLaMAOA, achiam2023gpt} and reasoning models (LRMs) \citep{guo2025deepseek, tian2025skipkv} have enabled strong general problem solving capabilities to artificial intelligence (AI) models spanning both discriminative (e.g., multiple-choice question answering~\citep{hendrycks2021measuring}) and generative (e.g., code generation\citep{liu2023your}) tasks.
%define models, solvers and experts are the same

% \RK{Since we are not doing traditional online LLM serving, can using the term serving to describe the work be confusing/misleading?}
Large language models (LLMs) \citep{touvron2023llama} yielding state-of-the-art (SoTA) accuracy on certain category of tasks, often fail to provide reasonable performance on different domains 
% (\textit{out-of-distribution}), 
due to their inherent training set and architectural bias towards certain tasks.
Skill based adoption of LLMs \citep{chen2024reconcile, wang2025slm, wang2025mixture, du2024improving, wang2024fusing} has significantly boosted their performance across a diverse set of tasks, including mathematical reasoning \citep{yu2024metamath}, commonsense and medical reasoning \citep{lobo2025impact}.
In particular, recent works demonstrate the benefits of \textit{mixture-of-agents} (MoA) frameworks, which harness the collective expertise of multiple LLMs to yield a unified robust system
that effectively bridges out-of-distribution generalization gaps \citep{azizi2026power, chu2501rl}.
In these frameworks, the task-based expert models
% \footnote{We alternatively refer a monolithic expert agent as solver or model in this paper.} 
% \footnote{We use \emph{expert} throughout for a member LLM in the MoA.}
% \TODO{replace all solvers with expert}
are meticulously chosen to construct an ensemble for the test-time mixture.
%
% However, they often suffer from iterative generation from the \textit{proposer models} \citep{wang2025slm} that often follows an \textit{aggregation} overhead \citep{wang2025mixture}.

% To improve the skill-based decision making of MoA, Skill-MoE \citep{chen2025symbolic} proposes adaptive, instance-level, skill-based routing.
%
% Additionally, it demonstrates the effectiveness of task-specific aggregator selection, avoiding the hard selection of best proposing model as aggregator.
%
% The skill-based router performs proposer and aggregator selection from a pool of LLMs and large reasoning models (LRMs) \citep{tian2025skipkv},  which produce much longer, verbose outputs. 
%
% These highly capable reasoning models are trained on long reasoning trajectories and naturally produce much longer, verbose outputs. These are often preferred by the router.
%

However, while MoA systems significantly improve accuracy, they introduce severe computational and latency overheads during inference. Typically, these frameworks rely on a two-stage process: an initial generation phase where multiple candidate experts independently process the query, followed by an aggregation phase where a final model synthesizes these diverse outputs into a cohesive response \citep{wang2025mixture, chen2024reconcile}. Recent advancements, such as Skill-MoE \citep{chen2025symbolic}, attempt to optimize this via adaptive, instance-level routing across a diverse pool of standard LLMs and large reasoning models (LRMs) \citep{tian2025skipkv}. LRMs naturally produce heavily verbose reasoning trajectories, drastically increasing the overall token count. Consequently, inference time suffers from sequential model loading delays, high synchronization costs in multi-GPU setups, and massive prefill overheads when the aggregator model must process the concatenated context of all upstream experts.
% However, in the expert phase, this generates a higher token count than other MoA baselines \citep{wang2025mixture}\citep{chen2024reconcile}.
%
% This is due to its preference for recruiting highly capable reasoning models, which are trained on long reasoning trajectories and naturally produce much longer, verbose outputs.
%
% Even though they introduce a batched inference strategy, inference time suffers from sequential model loading time for all experts and high synchronization cost when deployed in multi-GPU setup.
%
% Additionally, as the aggregator LLM must process the combined output or reasoning traces of its experts, the prefill of aggregation can be extremely lengthy and computationally expensive at times.
%

Motivated by this gap in MoA serving, we present MOSAIC, a workload-aware MoA scheduling framework designed for offline, throughput-oriented scenarios where a batch of queries and their routed experts are known in advance. This scenario is highly representative of modern LLM workloads, including benchmark evaluations, bulk annotation, synthetic-data generation, LLM-as-judge pipelines~\citep{zheng2023judging}, and commercial batch inference APIs~\citep{anthropic_batch_api, openai_batch_api}.
% \SM{eliminate if already evident}Online, SLO-driven multi-LLM serving~\citep{10.5555/3692070.3692543} is complementary and out of scope.
% \SK{Mention the limitations of symbolic MoE in 2-3 setences max}.
In this context, the performance of batched queries is determined by two key metrics: the overall makespan of the batch, and the time each query's answer becomes available (time-to-answer, TTA). Our contribution are as follows:
%
% The time-to-answer is governed by the structure of MoA: a query is answered only after enough of its experts have answered, and its experts run in different sessions on different GPUs.}\SM{Is the last sentence adding any value?}

\noindent
% \textbf{Our Contributions.}
%
% In specific, we make the following contributions:
%
% \begin{itemize}
\noindent \textbf{We characterize heterogeneous MoA workloads} on limited compute resources and systematically identify the bottlenecks. We highlight how combining instruction-tuned LLMs with long-reasoning models causes extreme variability in generation lengths (up to 13$\times$) and how skill-based routing induces popularity skew across expert pools (up to 92$\times$).

\noindent \textbf{We propose two-stage workload-aware scheduling.} At the expert placement stage, an ILP algorithm uses per-expert costs to decide model-to-worker placement, per-worker query counts and selective replication of heavy reasoning experts, minimizing the makespan. At the query-sequencing stage, MOSAIC orders each GPU's sessions and each session's prompts using per-query output-length predictions from the experts' own hidden states,
% trained on the same calibration set,
minimizing TTA.
% time-to-answer.
% \textbf{We formulate the expert-generation
% phase as an ILP} that jointly optimizes model-to-worker placement, per-worker prompt counts, and selective replication of heavy reasoning experts under a per-worker capacity limit.

\noindent 
\textbf{We introduce a confidence-aware adaptive aggregation gate} that bypasses the aggregator LLM on questions where the experts converge on a majority answer.

\noindent 
% We evaluate MOSAIC on a 4-GPU server across MMLU-Pro~\citep{wang2024mmlu}, MedMCQA~\citep{pal2022medmcqa}, and GPQA~\citep{rein2024gpqa}. Against a round-robin baseline\SM{change}
% at the first-seen-order of queries
% , MOSAIC delivers up to 2.54$\times$\ expert-stage and 4.23$\times$ aggregator-stage speedup, translating into 1.71-2.34$\times$ end-to-end wall-clock reduction while matching baseline accuracy within $\pm 1$pp. 
% \TODO{remove: Against a cost-aware greedy (LPT) scheduler, the ILP shortens the makespan by up to $25\%$ under tighter GPU budgets and finer-grained scheduling, and it remains effective under calibration noise, burst-wise dispatch and mixed workloads.}
\textbf{We evaluate MOSAIC on a 4-GPU server} across MMLU-Pro, MedMCQA, and GPQA. Against a round-robin baseline at the first-seen-order of queries, MOSAIC delivers $2.7$--$3.4\times$ end-to-end wall-clock reduction and $3.9$--$7.8\times$ lower mean time-to-answer, with accuracy within 1pp of the baseline.

% \end{itemize}

% routing-induced difficulty skew

% \begin{tabular}{lrrrrrrr}
% \toprule
% Mode & Wall (s) & LBE & Straggler & Generate (s) & Vis. Load (s) & Idle (s) & Idle \% \\
% \midrule
% dp\_allR & 6431 & 0.88 & 0.74 & 20482 & 286 & 3586 & 14.24\% \\
% dp\_allnonR & 892 & 0.97 & 0.94 & 1392 & 340 & 284 & 9.50\% \\
% dp\_default & 2442 & 0.93 & 0.88 & 6872 & 331 & 983 & 10.72\% \\
% roundrobin\_allR & 7082 & 0.57 & 0.25 & 15797 & 91 & 12082 & 42.91\% \\
% roundrobin\_allnonR & 632 & 0.57 & 0.29 & 1022 & 101 & 1029 & 43.42\% \\
% roundrobin\_default & 2663 & 0.55 & 0.06 & 5492 & 104 & 4720 & 45.03\% \\
% \bottomrule
% \end{tabular}

% \insertFigure{intro_ttft_4}{TTFT (a,b) and TPOT (c,d) scaling comparison of Qwen2.5-0.5B \citep{yang2025qwen3} and Mamba2-780m \citep{dao2024transformers}.  While Qwen is faster (1.9$\times$) at shorter sequence lengths, Mamba2's superior scaling provides a significant performance advantage(2.65$\times$) at longer contexts for both prefill and decode (for generation length 256 with batch size 1) stages.}

% \SM{define and refer model pool in the appendix}

% \input{Sections/02_Background}
\section{Background and Related Work}
\label{sec:background}
Recent work has shown that using multiple LLM responses can improve answer quality on challenging reasoning tasks.
Mixture-of-Agents (MoA) \citep{wang2025mixture} proposes a layered inference procedure in which multiple LLM agents generate responses, and later agents use previous responses as auxiliary context to produce a stronger final answer.
Skill-MoE \citep{chen2025symbolic} selects expert LLMs at the instance level using symbolic skill descriptions, then aggregating their generated reasoning traces into a final answer.

% \SR{can be removed if needed:
We focus on the scheduling problem that arises after expert selection: once the router assigns a subset of experts to each query, the system must execute these independent model calls efficiently on limited GPU resources.
% We focus on the scheduling problem created by such multi-agent inference workloads.
% %
% \SM{repetition? cut down here?}We formulate it as an offline scheduling task where a batch of questions and their assigned experts are known in advance.
% %
% Specifically, our problem begins after the expert selection step: once the skill-based router assigns a subset of experts to a question, the system must efficiently execute these independent model calls on limited GPU resources.
% %
% Such workloads naturally arise when evaluating multi-expert reasoning systems over fixed benchmark suites; for example, Skill-MoE \citep{chen2025symbolic} uses MMLU-Pro \citep{wang2024mmlu}, GPQA \citep{rein2024gpqa}, and MedMCQA \citep{pal2022medmcqa}.
%
%In this context, \emph{expert}, \emph{model}, and \emph{agent} refer to independent LLM instances, distinct from internal feed-forward layers in sparse MoE architectures or tool-calling agents.
% \SM{Is this paragraph redundant with the motivation start para}
This setting is different from standard LLM serving.
The pool of expert models can exceed available GPU capacity, and the skill-based router produces highly skewed selection frequencies across models.
Furthermore, mixing standard instruction-tuned LLMs with long-reasoning models introduces highly variable generation lengths.
These dynamics complicate resource allocation.
% }

\begin{figure}[t]
    \centering
    \includegraphics[width=\linewidth]{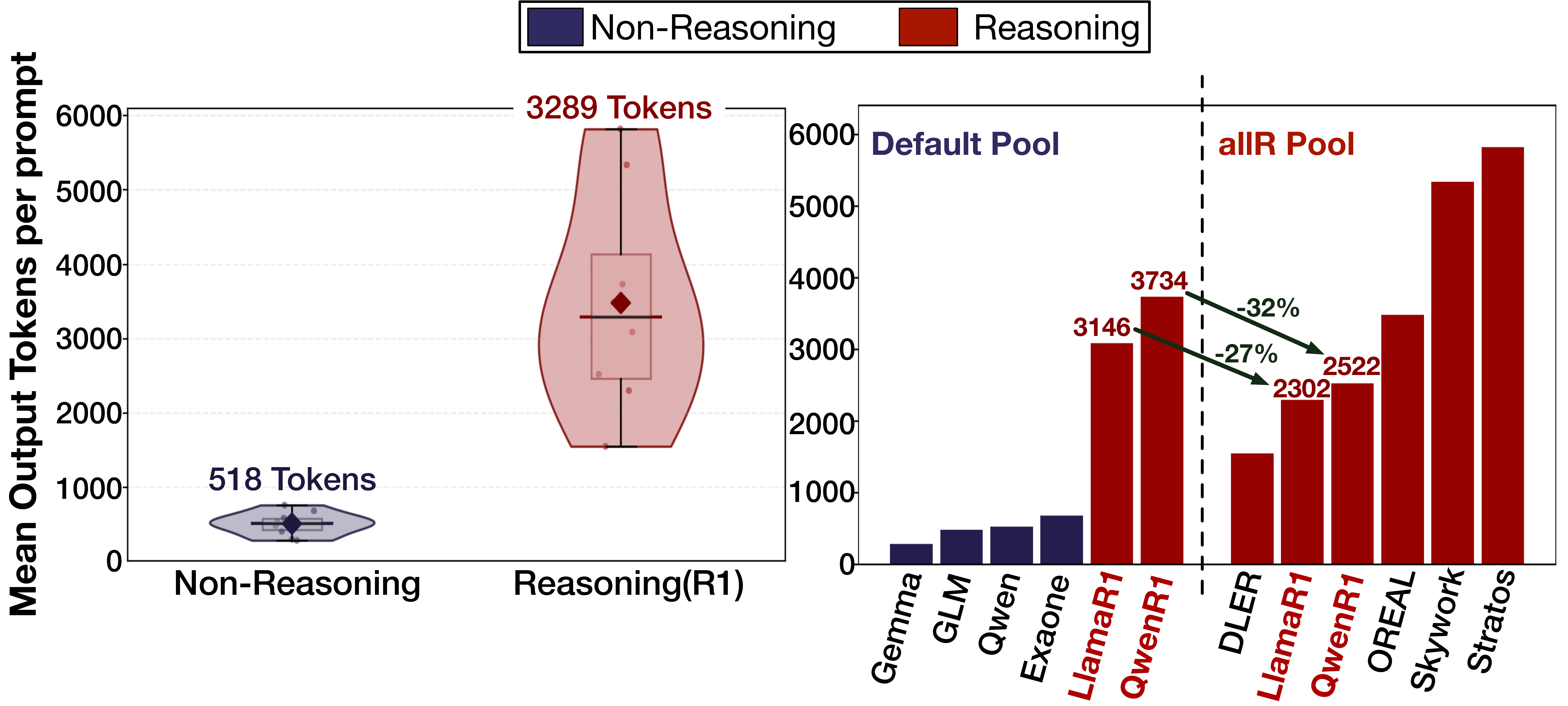}
    \vspace{-7mm}
    \caption{Output token distribution across non-reasoning and reasoning models (left), and mean output tokens per prompt across default and all reasoning (allR) expert pools (right).
    We show significant variability
    across experts and expert pools, with both R1 models producing fewer tokens in the allR pool.}
    \vspace{-4mm}
    \label{fig:tokenskew2}
\end{figure}

% \insertFigure{tokenskew2} {
% Output token distribution across non-reasoning and reasoning models (left), and mean output tokens per prompt across default and all reasoning (allR) expert pools (right).
% % The mean output tokens per prompt across non-reasoning and reasoning models (left), and across default and all reasoning (allR) expert pools (right).
% We show significant variability
% % in generated output tokens per prompt 
% across experts and expert pools, with both R1 models producing fewer tokens in the allR pool.
% %Output token distribution (a); and  mean output tokens per prompt (b) across reasoning and non-reasoning models for MMLU pro.
% }

\begin{figure*}[h]
    \centering
    \includegraphics[width=1\textwidth]{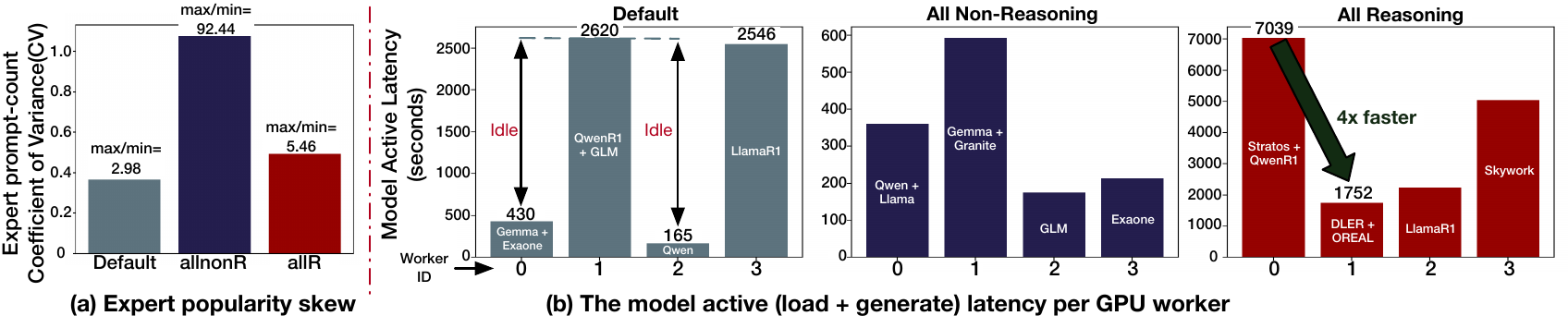}
    \caption{ (a) Expert popularity skew across model pools; (b) Overall state of GPU idling for MMLU-Pro}
    \vspace{-4mm}
    \label{fig:popularity2-shrinked}
\end{figure*}
% \insertWideFigureShrink{popularity2-shrinked}{(a) Expert popularity skew across model pools; (b) Overall state of GPU idling for MMLU-Pro}{1}

A natural approach to serving such workloads is \emph{static expert-per-worker placement}: deploy one high-throughput LLM serving instance per GPU and pin each expert to a fixed worker.
This follows the single-model serving regime optimized by systems such as vLLM and SGLang \citep{kwon2023efficient, 10.5555/3737916.3739916}.
While this avoids redundant model loading and preserves large per-expert batches, router-induced skew in expert demand and variability in generation length can leave some workers idle and others as stragglers.
On the other hand, \emph{data-parallel execution} can mitigate load imbalance by placing popular models on multiple workers, a strategy explored in traditional DNN serving systems such as INFaaS \citep{romero2021infaas}.
While DNN workloads often have predictable execution costs, LLM workloads differ fundamentally: generation is autoregressive, output lengths are input-dependent, and decoding cost can vary substantially across prompts and models.
Recent multi-LLM serving systems such as MuxServe \citep{10.5555/3692070.3692543} exploit LLM-specific prefill/decode behavior to colocate and multiplex multiple LLMs on a fixed GPU cluster.
However, they target online serving with SLO-oriented objectives, whereas we consider offline scheduling with a pre-routed batch of expert calls.
Online systems that buffer requests before dispatch~\citep{crankshaw2017clipper, nvidia_triton, yu2022orca} expose the same structure within each dispatch window;
Appendix~\ref{appendix:dynamic} shows that MOSAIC remains effective when it re-solves the schedule per window.
% \autoref{sec:eval:dynamic}\SM{point to appendix if 5.3 moved} shows that MOSAIC remains effective when it re-solves the schedule per window.
% \YZ{Length prediction for shortest-job-first is well studied for single-model serving: proxy predictors drive speculative SJF~\citep{qiu2024ssjf}, learning to rank shows that the ranking determines the gain~\citep{fu2024ltr}, and the served model's own embeddings predict lengths best~\citep{shahout2025trail}.}\SM{previous lines are important but abrupt}
%
% However, they target online serving settings with SLO-oriented objectives, whereas our setting is a finite offline batch of pre-selected expert calls.
% %
% \TODO{Finite batches also arise inside online systems that buffer requests into micro-batches prior to dispatch~\citep{crankshaw2017clipper, nvidia_triton, yu2022orca}; \autoref{sec:eval:dynamic} shows MOSAIC remains effective when it re-solves per burst.}
%
% Finally, while \emph{tensor/model parallelism} is critical for serving models that exceed single-GPU capacity \citep{li2023alpaserve, aminabadi2022deepspeed}, we assume each expert fits within a single GPU.
%
Our setting thus poses two coupled problems: placing experts across workers, and ordering the queries each worker executes.
For the latter, output-length prediction enables shortest-job-first scheduling in single-model serving~\citep{qiu2024ssjf, fu2024ltr, shahout2025trail}; MOSAIC extends it to multi-expert inference (\autoref{sec:qsched}).
%
% Our setting therefore poses two coupled problems: placing independent experts across a worker cluster, and ordering the queries each worker executes.
% For the latter, output-length prediction is well studied in single-model serving, where it enables shortest-job-first scheduling: a proxy model's predicted lengths drive speculative SJF~\citep{qiu2024ssjf}, learning to rank shows that a correct ordering, rather than exact lengths, determines the gain~\citep{fu2024ltr}, and the served model's own embeddings predict lengths more accurately than a separate proxy~\citep{shahout2025trail}.
% MOSAIC brings this ranking view to multi-expert inference, where each expert ranks its own prompts and the predictions order both sessions and prompts on every worker (\autoref{sec:qsched}).
%
% Therefore, the core challenge in our setting is the optimal placement and scheduling of independent experts across a worker cluster.

% \input{Sections/03_Motivation}

\section{Motivation}
\label{sec:motivation}

Serving multi-agent workloads presents a unique set of scheduling challenges.
% Given a batch of requests with pre-assigned experts, our objective is to minimize the overall makespan, each query's TTA and maximize hardware utilization.
We examine the distinct characteristics of these workloads below:

\betterparagraph{Observation 1}\textit{Heterogeneous model categories and expert pools yield highly diverse generation token count, with LRMs often showing long tail distribution of token count across candidate experts.}

\noindent
\autoref{fig:tokenskew2}(a) summarizes an experiment with MMLU-Pro, across 12 models of different classes (6 reasoning and 6 non-reasoning).
The LRMs' average generation length (3289 tokens/prompt) is $6.3\times$ longer than that of an LLM (518 tokens/prompt).
Further, LRMs often show long tail distribution of token generation, ranging from 1500 to 6000 tokens per prompt.
%(in \autoref{fig:tokenskew2}(a) it ranges from 1500 to 6000).
%
Additionally, the token-budget profile of an expert is not static but dependent on the composition of the broader expert pool.
\autoref{fig:tokenskew2}b shows two different expert pools, the default has two LRMs and four LLMs while the all reasoning (allR) pool has all six expert candidates as LRMs.
In default pool, LLamaR1 and QwenR1 being the only LRMs, generate 3146 and 3734 tokens per prompt on average respectively.
But on allR, they get a broader, mixed-difficulty subset of questions where their specific profile excels, with their token costs per prompt dropping to 2302 (-27\%) and 2522 (-32\%).
Meanwhile, the most-demanding questions migrate to specialized long-CoT experts (Skywork: 5339 tokens/prompt; Stratos: 5822 tokens/prompt).
\textbf{This shows that a model’s compute profile is dynamically shaped by the expert pool, since it affects the difficulty of the questions routed to that model}.

% \betterparagraph{II. Solver Popularity Skew} 
% \SK{follow earlier style and make it precise and to the point.}

\betterparagraph{Observation 2}
\textit{Popularity of different experts across model pools largely vary as a result of skill-weighted sampling by the router.}
\noindent
% Another major contributor to workload imbalance is the uneven popularity of solvers under skill-weighted sampling from the model pool.
%

\noindent
With the default reasoning + non-reasoning mix on the MMLU-Pro test set (2100 questions $\times$ 3 experts = 6300 selections), Gemma is selected most often (1632) and Exaone least (547), giving a max/min ratio of 2.98$\times$.
In \autoref{fig:popularity2-shrinked}a, y-axis represents population coefficient-of-variance for each model pool.
In all non-Reasonings (allnonR), the strongest remaining models dominate (Gemma 53\%, Qwen 24\%) while Llama and Granite each fall below 2\%, pushing max/min to 92.4$\times$.
\textbf{This shows that the popularity skew of experts is sensitive to the pool composition.}

\autoref{fig:popularity2-shrinked}b shows what happens when these skewed expert demands combine with token-budget disparity.
We consider a round-robin allocation (in first-seen-expert order in the available query queue), a scheme that commits each expert to a worker before any cost or count signal is observed.
In the default pool, reasoning workers dominate wall time: worker 1 (QwenR1+GLM, 2620s) and worker 3 (LlamaR1, 2546s) take 6-16$\times$ longer than worker 0 (Gemma+Exaone, 430s) and worker 2 (Qwen, 165s).
Notably, worker 0 (W0) holds the most popular expert (Gemma) in the entire pool, yet finishes earlier; because reasoning models generate 6-13$\times$ more tokens per prompt.
Per-prompt cost dominates popularity-count whenever the two expert classes mix.
The idle GPU time is also significant for allR and allnonR cases.
%
% W0 of allR got two reasoning solvers (Stratos+QwenR1) and spent 7039s active; where W1 (DLER+OREAL) and W2 (LlamaR1) finishes 4$\times$ and 3.15$\times$ earlier respectively.
%
% Each of W1 and W2 process same number of prompts (25.4\% and 25.1\%), but DLER prompt-cost being cheaper, it finishes faster.
%
No single factor explains the imbalance in GPU idling: a cost-aware scheduler must balance $\sum(\text{prompts} \times \text{per-prompt cost})$ per worker rather than either quantity alone.

\begin{figure}[t]
    \centering
    \includegraphics[width=\linewidth]{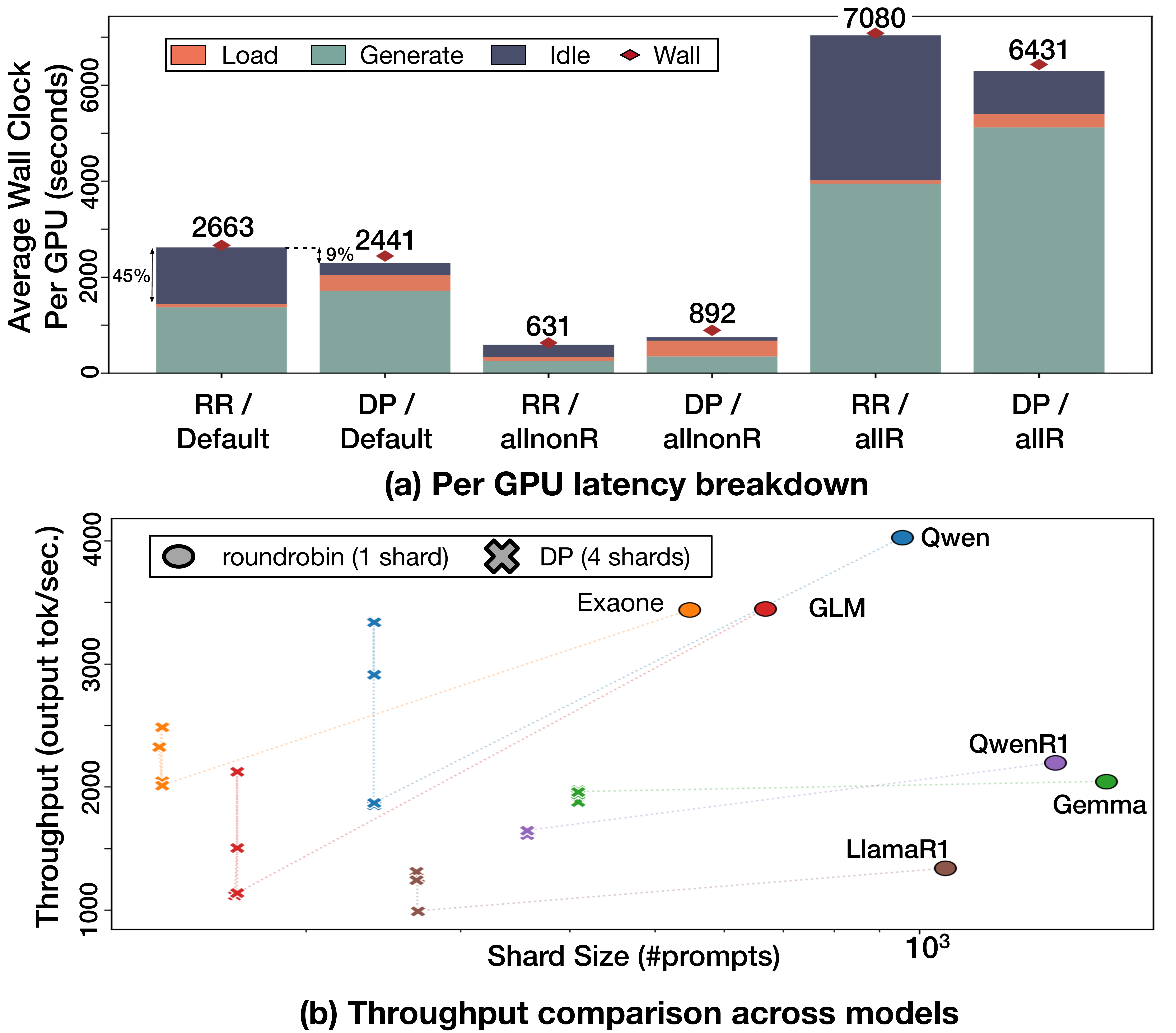}
    \vspace{-7mm}
    \caption{Comparing data-concurrent replication with round-robin allocation}
    \vspace{-4mm}
    \label{fig:dp-shrinked}
\end{figure}
% \insertFigure{dp-shrinked}{Comparing data-concurrent replication with round-robin allocation}
\betterparagraph{Observation 3}
\noindent
\textit{Data-concurrent replication trades straggler-idle time for two new costs: serialized model loads and per-shard throughput collapse.}
\noindent
Data parallelism (DP) can counter the previous GPU-idling by keeping a complete copy of each expert on all GPUs and shard its prompts evenly across workers.
\autoref{fig:dp-shrinked}a confirms the tradeoff is pool-dependent.
DP cuts the \textasciitilde45\% idle share of roundrobin runtime to give a overall \textasciitilde9\% wall-time improvement in default model-pool, and similarly improves allR (7080s to 6431s).
In allnonR, however, DP is \textasciitilde29\% slower (631s vs 892s), the per-expert generate budget is too small to amortize serialized loads.
DP also incurs a second cost from over-sharding: once each model's prompts are split four ways, per-shard throughput drops 40–60\% (in \autoref{fig:dp-shrinked}b) for the smaller instruction-tuned experts (Qwen, Exaone, GLM).
\textbf{Taken together, DP only wins when each expert's generate budget is large enough to absorb both the serialized load tax and the per-shard throughput loss}.
Full per-expert breakdown and critical-path analysis are in \autoref{appendix:dp_shard}.

\betterparagraph{Observation 4}
\noindent
\textit{Beyond placement, the order in which queries run determines when answers become available.}
\noindent
The order of expert sessions on a GPU leaves the makespan unchanged but shifts when questions are answered: running the short instruction-tuned model's session earlier cuts the median time to a question's last expert answer from 517\,s to 335\,s at the same 1490\,s makespan (\autoref{tab:session-order}).
Most questions are also decided early: $84.6\%$ of MMLU-Pro questions have two agreeing experts, whose second answer arrives at a median of 260\,s versus 465\,s for all three recruited experts.
Prompt order within a session can be optimized too, since an expert's output lengths can be ranked before generation.
% (Spearman $0.72$--$0.73$ for reasoning experts; \autoref{appendix:predictor}).
\textbf{A scheduler therefore can have a second level of decisions: the order of sessions and prompts, and when to stop waiting for experts whose votes no longer matter.}

% % \input{Sections/03_Benchmark}
\section{MOSAIC: Methodology}

\begin{figure*}[h]
    \centering
    \includegraphics[width=1\textwidth]{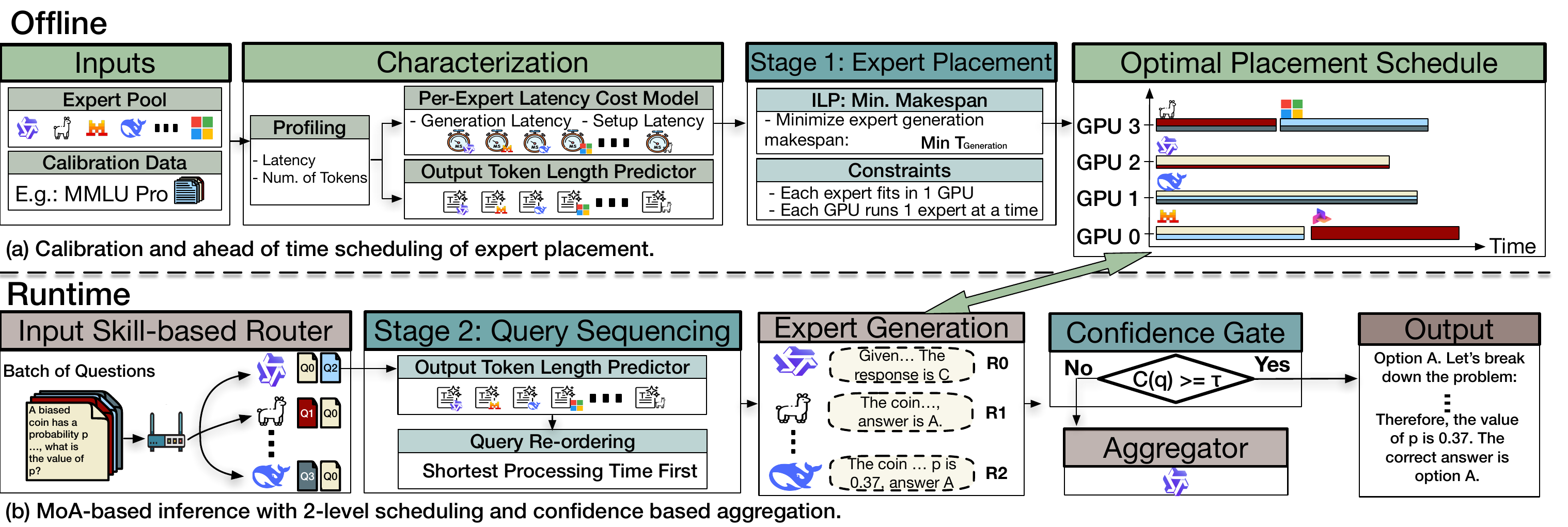}
    \caption{ Overview of the proposed MOSAIC framework: (a) The offline phase profiles expert latency and utilizes an ILP to determine an optimal expert placement schedule across GPUs. (b) The runtime phase features skill-based routing, query sequencing based on shortest predicted processing time, and confidence-gated adaptive aggregation.}
    \vspace{-4mm}
    \label{fig:overview-shrinked-v2}
\end{figure*}

\autoref{fig:overview-shrinked-v2} illustrates the proposed MOSAIC framework. The input prompts are routed to specific experts based on skill-based selection, and then executed on a multi-GPU server to generate the expert outputs. We aggregate all expert outputs to generate the final output. MOSAIC relies on two key components: (a) a two-stage workload-aware scheduling of the expert-generation phase, comprising an \textit{expert placement} stage to allocate models to workers and a \emph{query-sequencing} stage to decide the order of query execution; and (b) confidence-aware adaptive aggregation.

% \insertWideFigureShrink {overview-shrinked-v2} {
% % An overview of MOSAIC flow.
% % \SK{Use"MoA-based inference serving instead of "Symbolic MoE in caption of (b)", }
% Overview of the proposed MOSAIC framework: (a) The offline phase profiles expert latency and utilizes an ILP to determine an optimal expert placement schedule across GPUs. (b) The runtime phase features skill-based routing, query sequencing based on shortest predicted processing time, and confidence-gated adaptive aggregation.} {1}

\subsection{MOSAIC: Workload-aware Scheduling}

\subsubsection{Problem Setting}
\label{sec:c:method:setting}

% We consider an MoA deployment that serves a
% fixed prompt set on $N$ workers ($n \in \{1,\dots,N\}$) drawn from a
% pool 
% % (\autoref{tab:full_model_pool}) 
% of $M$ heterogeneous experts $\mathcal{M}$ with $M > N$. Each
% worker holds one model in GPU at once and executes its tasks
% as sequential vLLM sessions. For every expert $m$, let $\ell_m$ be the
% session-start (load) cost. The upstream skill-based recruit step is
% completed before scheduling and supplies the per-expert total prompt
% count $N_m$. Each worker's wall time decomposes into a setup term
% $\sum_m \ell_m \mathbf{1}[\text{worker } n \text{ loads } m]$ and a
% generation term; the scheduling objective is to minimise the makespan.
% \begin{equation}
% C \;=\; \max_{n \in \{1,\dots,N\}} T_n .
% \label{eq:c:makespan}
% \end{equation}
We consider an MoA deployment that serves a fixed prompt set across $N$ workers ($n \in \{1,\dots,N\}$) drawn from a pool of $M > N$ heterogeneous experts $\mathcal{M}$.
Each worker holds one model at a time and executes its tasks as sequential vLLM sessions; $\ell_m$ is the session-start (load) cost of expert $m$. We assume the upstream skill-based routing step is completed prior to scheduling, providing the total prompt count required from each expert.

A worker's wall time $T_n$ is the sum of its load and generation times.
% , and the first objective is the makespan $C = \max_{n} T_n$.
% [NAACL] second objective
% The second is the time-to-answer $A_q$ of every question $q$: the time from the start of the batch until its final answer is available.
%
MOSAIC optimizes for two primary objectives. The first objective is to minimize the makespan $C = \max_{n} T_n$. The second objective is to minimize the time-to-answer (TTA) $A_q$ for every question $q$, defined as the duration from the start of batch processing until its final aggregated answer is available.
% MOSAIC addresses the two objectives at two stages: the first stage decides which worker runs which experts and how many prompts each receives, which determines $C$. The second stage decides in which order the sessions of a worker and the prompts of a session run and when a question's answer is released, which determines $A_q$.

% \SM{placement level; question level do not sound good. replace last sentence 'None of the question-level decisions adds work, and one of them removes it'}

\subsubsection{Workload Characterization}
\label{sec:c:method:workload}

Output length is the dominant cost driver of MoA serving: once the batch has reached steady state, generation time grows linearly with output tokens, and each expert has a stable, model-specific output-token distribution (short responses for instruction-tuned experts, heavy long tails for reasoning-tuned ones).
We therefore run each expert once on a small calibration subset of the training data 
% and derive two artefacts from this run.
to derive two key metrics for scheduling.

\noindent\emph{Per-expert cost (expert placement).} From the expert's mean and maximum output-token counts
\begin{equation}
\bar{O}_m \;=\; \mathbb{E}[O \mid m],
\qquad
O_m^{\max} \;=\; \max_{q} O_{m,q},
\label{eq:c:per-model-output-stats}
\end{equation}
a per-model regression $T_m(N_m) = a_m N_m \bar{O}_m + b_m O_m^{\max}$ (batched decoding plus the drain of the longest sequence) is fitted from a short multi-$N$ profiling sweep on the calibration set, which predicts the session wall time as the prompt count $N_m$ varies across experts. Dividing $T_m$ by $N_m$ gives the per-prompt scheduling cost
\begin{equation}
\tau_m \;\triangleq\; \frac{T_m(N_m, \bar{O}_m, O_m^{\max})}{N_m},
\label{eq:c:tau}
\end{equation}
which is a constant coefficient at solve time.

\noindent\emph{Per-prompt length ranks (query sequencing).} Within an expert, the length of an individual output can be ranked: a ridge regressor on the expert's own hidden state of the last prompt token, fitted on the same calibration run as expert placement, orders the prompts of a session by expected output length (\autoref{appendix:predictor}).
% These ranks are used only to order work; they never enter the placement objective.

% \subsubsection{Scheduling Algorithm Formulation}
\subsubsection{Stage 1: Expert Placement}
\label{sec:c:method:ilp}

% In our mixture-of-agent setup, the scheduling decisions are discrete: model placement on a worker is binary, prompt counts are integer, and the capacity
% bounds are small integer caps. Therefore, the schedule is naturally cast as an ILP. A greedy
% list-scheduling heuristic would forfeit the joint optimisation of
% placement and per-worker prompt count under the replica cap; at the
% problem sizes encountered here ($M \le 6$ experts, $N = 4$ workers),
% the search space is small enough that a modern CP-SAT solver returns
% provably optimal solutions in milliseconds, so the price of exactness
% over a heuristic is negligible.
% \SM{last line might need replacement. following is just an example: Exact search matters most when the problem is far from fluid: with few workers per heavy expert, a first-order load cost per placement, or indivisible scheduling units (e.g., per-category prompt blocks), an early, locally balanced greedy placement cannot be repaired later. \autoref{sec:eval:greedy} quantifies this against a cost-aware greedy baseline.
% The ILP remains cheap: CP-SAT~\citep{ortools} solves our 6-expert/4-worker instances to optimality in under $0.1$\,s (\autoref{appendix:scaling}).}
Placement decisions are discrete (binary placements, integer prompt counts, small integer caps), so we cast placement as an ILP over the per-expert costs; at our problem sizes ($M \le 6$, $N = 4$) CP-SAT~\citep{ortools} solves it optimally in milliseconds, and the formulation does not depend on the solver (\autoref{appendix:greedy}).

\betterparagraph{\texorpdfstring{Decision variables and worker time.}{Decision variables and worker time.}}
For every $(m, n)$, $Y_{m,n} \in \{0,1\}$ indicates that
worker $n$ loads model $m$, and $X_{m,n} \in \mathbb{Z}_{\ge 0}$
counts the model-$m$ prompts placed on worker $n$:
\begin{equation}
T_n \;=\; \sum_{m} \ell_m \, Y_{m,n}
       \;+\; \sum_{m} \tau_m \, X_{m,n}.
\label{eq:c:worker-time}
\end{equation}

\betterparagraph{Constraints.}
Every prompt of every model must be placed on some worker, and prompts of model $m$ may only be served by workers that load $m$:
\begin{equation}
\sum_{n} X_{m,n} \;=\; N_m, \qquad X_{m,n} \;\le\; N_m \, Y_{m,n}, \quad \forall m, n.
\label{eq:c:coverage}
\end{equation}
Each model is loaded on at least one worker and at most on a per-model cap $R_m^{\max}$, and no worker hosts more than $K$ distinct models:
\begin{equation}
\begin{gathered}
1 \;\le\; \textstyle\sum_{n} Y_{m,n} \;\le\; R_m^{\max} \quad \forall m, \\
\textstyle\sum_{m} Y_{m,n} \;\le\; K \quad \forall n.
\end{gathered}
\label{eq:c:replica}
\end{equation}
The cap admits a model to as many workers as its execution time can amortise one additional load each. Writing $\hat{T}_m \triangleq T_m(N_m, \bar{O}_m, O_m^{\max})$, we define:
\begin{equation}
R_m^{\max} \;=\; \mathrm{clamp}\!\left(1,\;
   \left\lfloor \tfrac{\hat{T}_m}{\ell_m} \right\rfloor,\;
   N\right),
\label{eq:c:rho-replica-cap}
\end{equation}
Consequently, reasoning experts are eligible for replication up to $N$ workers, whereas models with comparable load and execution costs are pinned to a single worker. With $K < M$ the solver cannot place every expert on its own worker, which forces the longest-running experts to be replicated.

\betterparagraph{Objective.}
$\min C$ subject to $C \ge T_n$ for all $n$.
Integer prompt counts permit asymmetric splits of an expert across workers, and the resulting topology, in which each reasoning replica is paired with a distinct lightweight partner, emerges from the joint minimization. 
% rather than being designed in; a replicated expert's prompts are interleaved across its replicas so that each receives a representative mix.
Furthermore, a replicated expert's prompts are interleaved across its replicas to ensure each worker processes a representative complexity mix.

% [NAACL] new subsection
\subsubsection{\texorpdfstring{Stage 2: Query Sequencing}{Stage 2: Query Sequencing}}
\label{sec:qsched}
Expert placement fixes the sessions of each worker and the prompts of each session, but not when individual questions are answered: a question can be answered only after enough of its experts have responded, and these experts run in different sessions on different workers.
The query-sequencing stage orders this fixed workload by predicted output length. Since it only reorders execution, the makespan of the placement is preserved.

\betterparagraph{\texorpdfstring{Inter-session order}{Session order.}}
Each worker runs its sessions in increasing order of predicted total output tokens; on MMLU-Pro, for example, each worker's non-reasoning session, which finishes within minutes, runs before its reasoning session.
The short answers that most questions need, and the gate decisions that depend on them, therefore arrive early (\autoref{tab:session-order}).

\betterparagraph{\texorpdfstring{Intra-session prompt order}{Prompt order.}}
A session generates outputs for a bounded number of prompts at a time while the remaining prompts wait in a queue; MOSAIC serves this queue in order of increasing predicted output length (shortest processing time first, SPT), which lowers the mean completion time of the session's prompts.
The predictor only needs to rank the prompts (\autoref{sec:c:method:workload}), and the order matters only for sessions with more prompts than fit into one batch: all sessions on MMLU-Pro, all but two on MedMCQA, and none on GPQA.

\subsection{MOSAIC: Adaptive Aggregation}
\label{sec:aggregation}
In expert selection phase, experts are recruited per-question against varied skill profiles.
The $k$ experts answering a question are deliberately diverse.
When such skill-aligned experts converge on the same answer, that convergence is a strong signal that the answer is correct.
It is potentially stronger than $k$ independent samples from a single model (self-consistency).
We exploit this signal to skip the aggregator LLM on questions where experts agree.

\betterparagraph{Observation} On MMLU-Pro, $960$ of $2100$ questions ($45.71\%$) get a unanimous $3{:}0$ vote among the three recruited experts.
In this $3{:}0$ bucket, the un-aggregated majority answer matches the golden output $82.92\%$ of the time.
It is essentially similar (or better) in accuracy when all queries in this bucket are fully passed to the aggregator LLM ($82.40\%$).
Similar trend holds on the other two benchmarks: $3{:}0$-bucket covers $47.48\%$ (MedMCQA),
% $43.33\%$ (AIME24)\SM{remove},
and $37.37\%$ (GPQA) of questions, with majority accuracy matching or exceeding the aggregator's in every case (\autoref{tab:adaptive-agg}).

% \betterparagraph{Confidence-gated aggregation}
Let $\{a_1,\dots,a_k\}$ be the answers of $k$ recruited experts to a question $q$, and let $C(q)\in[0,1]$ be a confidence function over those answers.
We define a confidence gate with threshold $\tau$:
\begin{equation}
\hat{y}(q) \;=\;
\begin{cases}
\operatorname{majority}\!\bigl(\{a_i\}_{i=1}^{k}\bigr) & \text{if } C(q) \ge \tau, \\
\operatorname{Aggregator}\!\bigl(q, \{a_i\}_{i=1}^{k}\bigr) & \text{otherwise.}
\end{cases}
\label{eq:adaptive-agg}
\end{equation}
Here we instantiate $C$ as the \emph{plurality fraction}
\[
C_{\text{vote}}(q) \;=\; \max_{a}\;\frac{\bigl|\{i : a_i = a\}\bigr|}{k},
\]
So for $k{=}3$ the buckets $3{:}0$, $2{:}1$, and $1{:}1{:}1$ correspond to $C{=}1$, $C=\frac{2}{3}$, and $C=\frac{1}{3}$ respectively.
We report two operating points: a conservative gate $\tau{=}1$ (gate $3{:}0$ only) and an aggressive gate
$\tau{=}2/3$ (gate $3{:}0$ and $2{:}1$).
Algorithm is available in \autoref{appendix:agg_algo}.

%Decide which one to keep (algo or equations?)

%
The formulation in \autoref{eq:adaptive-agg} is agnostic to the choice of confidence estimator.
%
% Inter-expert agreement is one instantiation, but any signal correlated with response quality may be substituted into the same gate.
% Any signal correlated with response quality (for example response-level uncertainty, or context-aware features) can be substituted for inter-expert agreement in the same gate.
%
% We use the plurality fraction in this work because it is parameter-free, and---as we show in \autoref{sec:adaptive-results}--- is already strong enough to recover most of the available compute savings without compromising accuracy.
% The gate in \autoref{eq:adaptive-agg} is agnostic to the confidence estimator.
For open-ended outputs, where plurality is ill-defined, $C$ can be defined using normalized sequence entropy (per-token entropy over top-20 log-probabilities, averaged across the response and normalized by the expert's mean), invoking the aggregator when even the entropy of the most confident expert exceeds a threshold.
Correct responses have consistently lower entropy than incorrect ones, including on free-form AIME24 answers.
Furthermore, incorrect $2{:}1$ agreements show a larger entropy gap between the two agreeing experts than correct ones (\autoref{appendix:entropy}), enabling entropy to flag unreliable majorities.
% \SM{ Keep the wording as an analysis unless E4 (end-to-end entropy \% gate) is done; then add the measured accuracy and skip\% numbers.}

\section{Evaluation and Results}
\label{sec:results}
% \insertWideFigure{expert-gantt}{Per-GPU execution timeline of MOSAIC schedule (bottom row) and the round-robin baseline (top row) on 3 selected datasets: MMLU\_Pro (left), MedMCQA (middle), and GPQA (right). Each bar represents one GPU worker session. MOSAIC reduces the workload's makespan by assigning balanced pairs of reasoning and non-reasoning experts to the workers.} 
\begin{figure*}[t]
    \centering
    \includegraphics[width=\textwidth]{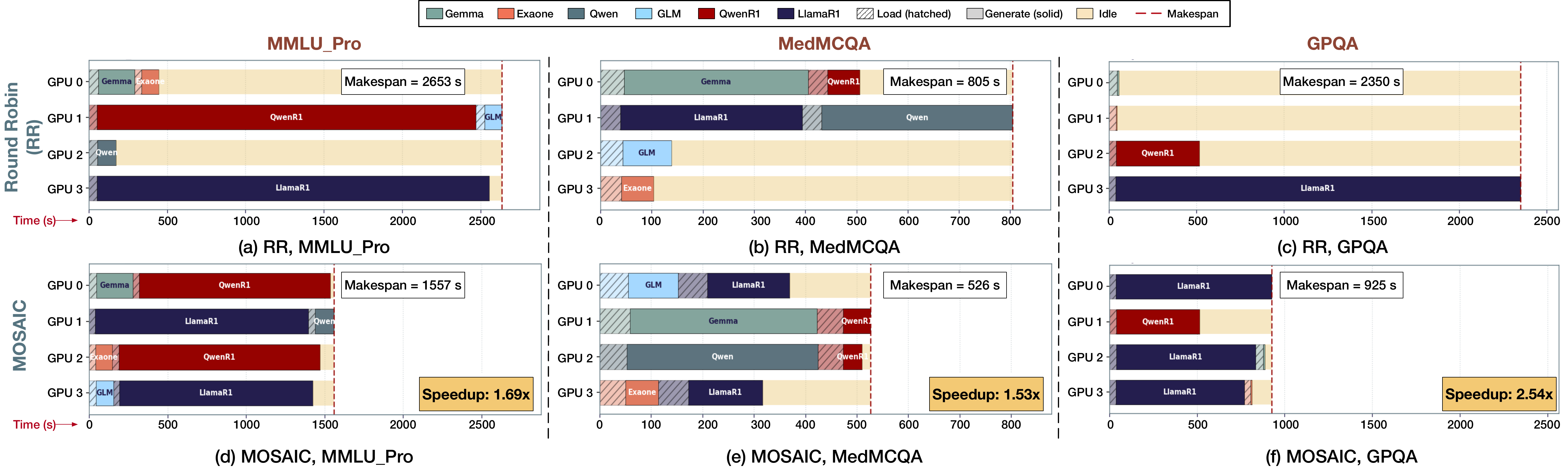}
    \vspace{-6mm}
    \caption{Expert-stage execution timelines per GPU for round-robin (RR; top: a--c) and MOSAIC, expert-placement optimization only with ILP (bottom: d--f) on MMLU-Pro, MedMCQA and GPQA. Each row is one GPU; each segment is one expert session, split into model load (hatched) and generation (solid). RR leaves GPUs idle behind one or two reasoning-expert stragglers, whereas MOSAIC reduces the makespan by $1.69\times$, $1.53\times$ and $2.54\times$.}
    \vspace{-4mm}
    \label{fig:expert-gantt}
\end{figure*}
% \insertWideFigure{expert-gantt}{Expert-stage execution timelines per GPU for round-robin (RR; top: a--c) and MOSAIC, expert-placement optimization only with ILP (bottom: d--f) on MMLU-Pro, MedMCQA and GPQA. Each row is one GPU; each segment is one expert session, split into model load (hatched) and generation (solid). RR leaves GPUs idle behind one or two reasoning-expert stragglers, whereas MOSAIC reduces the makespan by $1.69\times$, $1.53\times$ and $2.54\times$.}

\subsection{Evaluation Methodology}
\label{sec:eval:methodology}
\betterparagraph{Platform and datasets}
All experiments run on a single host equipped with four NVIDIA A100
80~GB GPUs, using vLLM 0.7.1. One model is loaded per GPU at a time and swapped
sequentially across vLLM sessions. The single-host, multi-GPU
configuration is the minimal setting that exposes the cross-worker
scheduling problem the paper formalises while excluding network-level
confounds; the 80~GB envelope per GPU is large enough to hold any one
of our experts together with its full 32\,k-token KV cache without
sharding, so the only setup cost incurred by the scheduler is the
per-session model load $\ell_m$. All three benchmarks are served from the
same hybrid experts pool (reasoning-tuned and instruction-tuned experts; full list in
Appendix~\ref{appendix:pool}); per-expert prompt slot counts on each
benchmark are reported in Appendix~\ref{appendix:datasets}.

\betterparagraph{Calibration and baseline}
The per-model output statistics $\bar O_m$, $O_m^{\max}$ are
estimated from a small train-only subset of each benchmark (350
questions for MMLU\_Pro and 504 for MedMCQA); the scheduler never
observes the test set. The resulting schedule is robust to the size, noise and source of the calibration set (Appendix~\ref{appendix:robust}).
The comparison baseline is round-robin (RR)
placement, $m \mapsto m \bmod N$, which requires no calibration and
incurs exactly one load per model. RR is the natural multi-GPU
baseline against which any non-trivial scheduling policy must
demonstrate value.
% \SM{replacement dummy: The per-model output statistics $\bar O_m$, $O_m^{\max}$ are estimated from a small train-only subset of each benchmark (350 questions for MMLU\_Pro and 504 for MedMCQA \SM{GPQA calibration source and size?}); the scheduler never observes the test set.
% We compare against two baselines.
% Round-robin (RR) placement, $m \mapsto m \bmod N$, needs no calibration and loads each model once.
% LPT is a cost-aware greedy list scheduler (longest processing time first): it sorts scheduling units by their calibrated cost and repeatedly assigns the longest remaining unit to the least-loaded worker \SM{define the unit for the coarse setting}.
% LPT consumes exactly the same calibration as MOSAIC, so their difference isolates the search strategy from cost-awareness.
% Wall times are means over three routing seeds with 95\% confidence intervals \SM{E1}.
% Replication does not raise per-GPU memory, since each worker holds one model at a time; its only cost is the additional load $\ell_m$, which the ILP accounts for.
% }

\subsection{MOSAIC Accuracy Result and Analysis}
\label{sec:aggregator_results}

\begin{table}[h]
\centering
\small
\vspace{-5mm}
\caption{Accuracy ($\Delta$pp)
% \SM{, with 95\% paired-bootstrap CI: E5}
and aggregator-call savings (skip\%) under confidence-gated
aggregation. 
% $\Delta$pp is the accuracy change vs. the full-aggregator baseline;skip\% is the fraction of questions where the aggregator is bypassed.
}
\label{tab:adaptive-agg}
% \begin{tabular}{lccccc}
% \toprule
% Task & Baseline & \multicolumn{2}{c}{$\tau{=}1$ (3{:}0)} & \multicolumn{2}{c}{$\tau{=}2/3$ (3{:}0 + 2{:}1)} \\
% \cmidrule(lr){3-4}\cmidrule(lr){5-6}
%  & acc.\ (\%) & $\Delta$pp & skip\% & $\Delta$pp & skip\% \\
% \midrule
% MMLU-Pro & 63.19 & $+0.24$ & 45.71 & $-0.62$ & 84.62 \\
% MedMCQA  & 59.69 & $-0.04$ & 47.48 & $+0.32$ & 90.18 \\
% % AIME24\SM{remove}   & 53.33 & $+3.34$ & 43.33 & $+16.67$ & 73.33 \\
% GPQA     & 56.57 & $+1.00$ & 37.37 & $+0.49$ & 87.37 \\
% \bottomrule
% \end{tabular}
\resizebox{\columnwidth}{!}{%
\begin{tabular}{lccccc}
\toprule
Task & Baseline & \multicolumn{2}{c}{$\tau=1$ (3:0)} & \multicolumn{2}{c}{$\tau=2/3$ (3:0 + 2:1)} \\
\cmidrule(lr){3-4}\cmidrule(lr){5-6}
& acc.\ (\%) & $\Delta$pp & skip\% & $\Delta$pp & skip\% \\
\midrule
% MMLU-Pro\citep{wang2024mmlu} & 63.19 & $+0.24$ & 45.71 & $-0.62$ & 84.62 \\
% MedMCQA\citep{pal2022medmcqa}  & 59.69 & $-0.04$ & 47.48 & $+0.32$ & 90.18 \\
% % AIME24\SM{remove} & 53.33 & $+3.34$ & 43.33 & $+16.67$ & 73.33 \\
% GPQA\citep{rein2024gpqa}     & 56.57 & $+1.00$ & 37.37 & $+0.49$ & 87.37 \\
MMLU-Pro & 63.19 & $+0.24$ & 45.71 & $-0.62$ & 84.62 \\
MedMCQA  & 59.69 & $-0.04$ & 47.48 & $+0.32$ & 90.18 \\
% AIME24\SM{remove} & 53.33 & $+3.34$ & 43.33 & $+16.67$ & 73.33 \\
GPQA     & 56.57 & $+1.00$ & 37.37 & $+0.49$ & 87.37 \\
\bottomrule
\vspace{-6mm}
\end{tabular}%
}

\end{table}

We evaluate the accuracy change for a task and the latency savings in the aggregation phase from the gating mechanism described in \autoref{eq:adaptive-agg}.
\autoref{tab:adaptive-agg} reports accuracy under the conservative ($\tau{=}1$, skip $3{:}0$) and aggressive ($\tau{=}2/3$, skip $3{:}0$ and $2{:}1$) gates against the full-aggregator baseline.
%
% The conservative gate matches the baseline within $1$pp on 
%all three large benchmarks (MMLU-Pro, MedMCQA and GPQA).
% and \emph{improves} accuracy on AIME24 and GPQA,
% large benchmarks like MMLU-Pro, MedMCQA and \emph{improves} accuracy on GPQA
% while bypassing $37$-$48\%$ of aggregator calls.
%
% The aggressive gate bypasses $73$-$90\%$ of aggregation calls while closely matching the baseline performance of all tasks considered.
%
The conservative gate stays within $1$pp of the baseline on all three benchmarks ($+0.24$, $-0.04$, $+1.00$pp) while bypassing $37$-$48\%$ of aggregator calls.
The aggressive gate bypasses $85$-$90\%$ of calls 
while closely matching the baseline performance of all tasks considered.
% with accuracy changes between $-0.62$ and $+0.49$pp.
% All speedups reported below use the aggressive gate.

\subsection{MOSAIC Speedup Result and Analysis}
% \subsubsection{Expert Scheduling Speedup}
\paragraph{Expert Scheduling Speedup}
\label{sec:eval:expert}

Figure~\ref{fig:expert-gantt} reports the expert-phase per-GPU
timelines of MOSAIC against the round-robin baseline on three
benchmarks. Each configuration is repeated over three random seeds to
control for the stochastic component of vLLM's continuous-batching
scheduler; per-seed walls are listed in
Appendix~\ref{appendix:multiseed}. MOSAIC delivers a $1.69\times$
speedup on MMLU\_Pro, $1.53\times$ on MedMCQA, and $2.54\times$ on
GPQA, with setup cost in every case comparable to RR's. The small
number of additional model loads required to replicate the heaviest
experts is paid back many times over by the resulting drop in
worker-wall imbalance. The mechanism is visible in the figure: in
every RR panel one or two workers dominate the wall, idling the
others (the dotted yellow regions); MOSAIC pulls the heaviest
expert onto two or more workers, pairs each replicated slice with a
single-worker instruction-tuned partner, and so contracts the
makespan to near the parallel lower bound. The same structural
decision drives all three speedups despite a $\sim$$20\times$
variation in workload size (594 prompts on GPQA, 12\,549 on MedMCQA)
and very different recruit distributions, indicating that the win is
not specific to any single benchmark's $\tau_m$ profile.

% \SM{the whole can be eliminated to save space. Added in appendix A.6}
% The capacity cap $K{=}2$ adopted throughout is the smallest value
% that satisfies the two requirements implied by
% \autoref{eq:c:rho-replica-cap}. First, a feasibility lower bound
% $K \ge \lceil M/N \rceil = 2$ holds independently of any calibration.
% Second, the per-model replica caps measured from calibration sit at
% $R_m^{\max}{=}N$ for the two reasoning experts ($\rho_m \approx
% 0.02$, since their generation time dwarfs the load cost) but
% $R_m^{\max}{=}1$ for the four instruction-tuned experts ($\rho_m \in
% [0.2, 0.95]$, where an extra load would consume most of the savings).
% $K{=}2$ is therefore precisely the capacity that grants the reasoning
% experts the room they need to split without admitting a second load
% of any instruction-tuned expert: $K{=}1$ is infeasible and $K{=}3$
% would over-replicate models whose $R_m^{\max}{=}1$ already forbids
% the extra load. The solver treats $K{=}2$ as a soft budget,
% saturating it on workers that benefit from a reasoning slice and
% leaving it slack elsewhere; this is most pronounced on GPQA, where
% the recruiter sends $72\%$ of all prompts to LlamaR1 and MOSAIC
% consequently replicates LlamaR1 across three workers, yielding
% the largest observed speedup, because RR's single-worker
% LlamaR1 allocation is the bottleneck.

We set the per-worker cap to $K{=}2$, the smallest feasible value ($K \ge \lceil M/N \rceil$); with the calibrated $\rho_m$ it lets the two reasoning experts ($R_m^{\max}{=}N$) split while pinning the four instruction-tuned ones ($R_m^{\max}{=}1$) (\autoref{appendix:kcap}).
On GPQA, where the recruiter sends $72\%$ of prompts to LlamaR1, MOSAIC replicates it on three workers, which gives the largest speedup.

% \begin{table}[t]
% \centering
% \small
% \caption{\SM{additional table. MedMCQA greedy cells need E2;
% % all cells  become mean +- CI after E1.
% If the length-prediction ILP is adopted, add
% a column "MOSAIC+LP" AND rerun Greedy with the same predictor (E3).}Expert-stage wall time (s) and speedup over RR. LPT uses the same calibration as MOSAIC.
% % ``cat.'' schedules whole (expert, category) blocks instead of divisible per-expert pools.
% }
% \label{tab:greedy}
% \resizebox{\columnwidth}{!}{%
% \begin{tabular}{lcrrr}
% \toprule
% Task & GPUs & RR & LPT greedy & MOSAIC (ILP) \\
% \midrule
% MMLU-Pro        & 4 & 2635 & 1678 (1.57$\times$) & \textbf{1557 (1.69$\times$)} \\
% MMLU-Pro        & 3 & 2779 & 2577 (1.08$\times$) & \textbf{2409 (1.15$\times$)} \\
% % MMLU-Pro (cat.) & 4 & 2635 & 2047 (1.29$\times$) & \textbf{1539 (1.71$\times$)} \\
% MedMCQA         & 4 & \SM{805} & \SM{E2} & \textbf{\SM{527} (1.53$\times$)} \\
% GPQA            & 4 & 2348 &  946 (2.48$\times$) & \textbf{935 (2.51$\times$)} \\
% GPQA            & 3 & 2340 & 1768 (1.32$\times$) & \textbf{1330 (1.76$\times$)} \\
% \bottomrule
% \end{tabular}%
% }
% \end{table}

A cost-aware greedy scheduler (LPT) given the same calibration recovers most of this gain at 4 GPUs ($1.57\times$/$2.48\times$ vs.\ $1.69\times$/$2.51\times$ for MOSAIC on MMLU-Pro/GPQA), but falls short when GPUs are scarce: with 3 GPUs MOSAIC reaches $1.76\times$ vs.\ $1.32\times$ on GPQA, and its advantage grows with fleet size (Appendix~\ref{appendix:greedy}).

\betterparagraph{Adaptive Aggregation Speedup}
\label{sec:eval:aggregator}
On MMLU-Pro, gating both the $3{:}0$ and $2{:}1$ buckets with the aggressive gate ($\tau{=}2/3$), reduces the aggregator latency from $1514.7\text{s}$ to $870.5\text{s}$, when we ran the aggregator with tensor-parallelism across 4 available GPUs.
This results in a $\mathbf{1.74\times}$ aggregator-stage speedup.
On MedMCQA, the same gate bypasses \textasciitilde$90\%$ of questions and yields $\mathbf{4.23\times}$ speedup ($516.49\text{s}$ to $122.09\text{s}$).
For GPQA, skipping \textasciitilde$87\%$ questions give speedup of $\mathbf{1.53\times}$ ($348\text{s}$ to $227.73\text{s}$).
Averaged over the three benchmarks, the aggregator stage is $2.5\times$ faster.
% The speedup is sub-linear in the skip rate because the remaining $1{:}1{:}1$ questions carry the longest expert traces ($2.53\times$ longer than $3{:}0$ on MMLU-Pro; Appendix~\ref{appendix:agg_tokens}).

\begin{figure}[t]
    \centering
    \includegraphics[width=\linewidth]{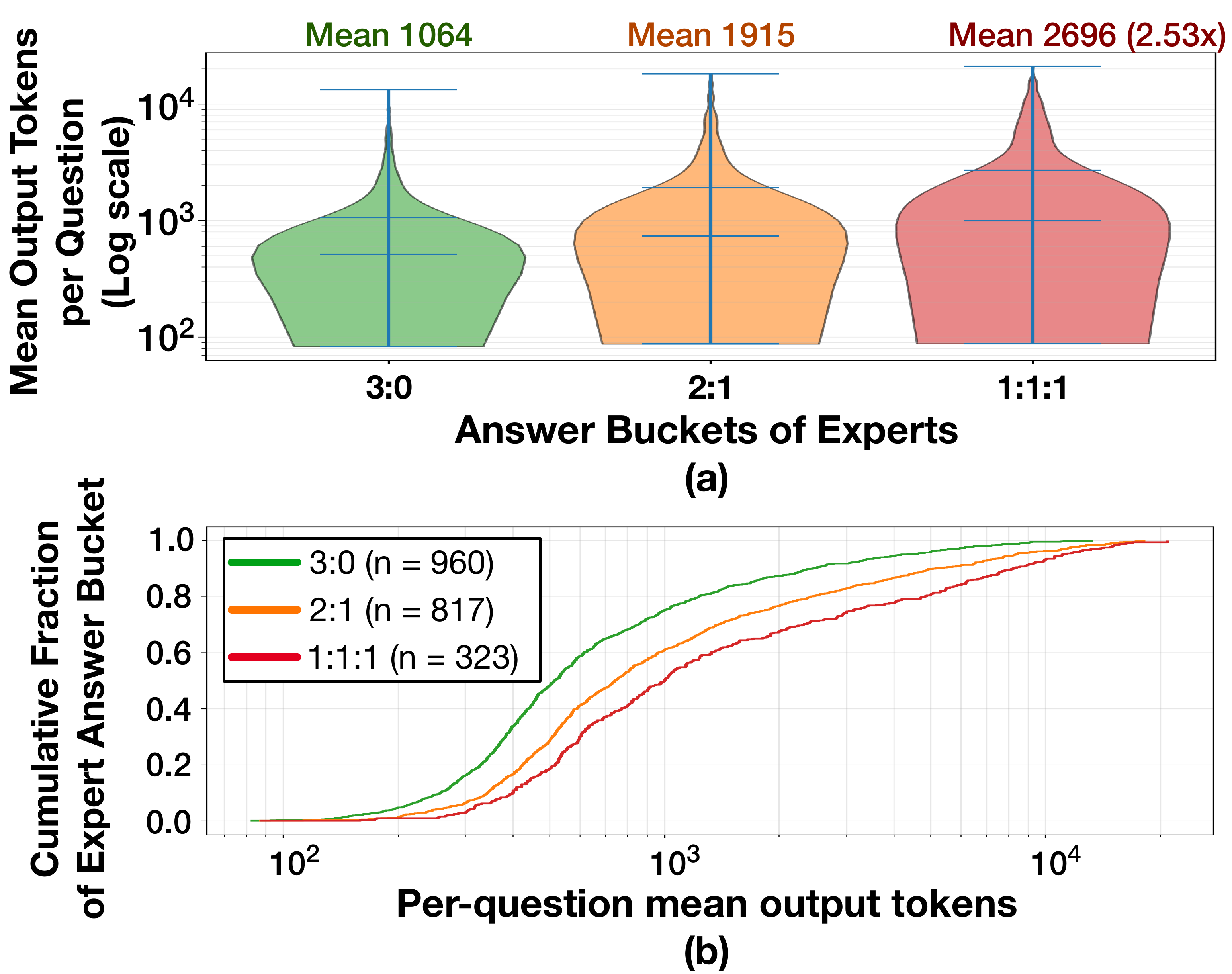}
    \vspace{-7mm}
    \caption{Per-question mean output tokens by agreement bucket. a: violin plot on a log scale, all experts pooled. b: ECDF by bucket. Unanimous ($3{:}0$) questions resolve in markedly fewer tokens than disagreement ($1{:}1{:}1$) questions ($2.53\times$ ratio, Spearman $\rho = -0.27$).}
    \vspace{-4mm}
    \label{fig:aggr_token}
\end{figure}

% \subsection{Ablation on aggregation skip results}
Speedup in our experiments is sub-linear to the gating rate in aggregator calls.
We bypass $84.62\%$ of MMLU-Pro questions but recover only $1.74\times$ rather than about \textasciitilde$6\times$.
The reason is that the residual $1{:}1{:}1$ bucket has the longest reasoning traces from the experts.
For MMLU-Pro, the mean output length is $2696$ tokens for $1{:}1{:}1$ vs.\ $1064$ tokens for $3{:}0$, a $2.53\times$ ratio.
The gating mechanism ensures that hard questions requiring expert-generated long traces are handled by the aggregator.
\autoref{fig:aggr_token}a shows the relationship between CoT length and expert agreement.
The mean output tokens per question decrease monotonically across agreement buckets and within each expert class.
The ECDFs (in \autoref{fig:aggr_token}b) show that questions having unanimous answer resolve in shorter traces well before the ones having most disagreement do (Spearman $\rho = -0.27$, $p < 10^{-30}$).
%
% The monotone relationship between bucket and trace length holds for both reasoning and non-reasoning solvers independently (Fig.~\ref{fig:tokens-vs-bucket}), ruling out a confound where the effect is driven by a single solver class.
% \insertFigure{aggr_token}{Per-question mean output tokens by agreement bucket. a: violin plot on a log scale, all experts pooled. b: ECDF by bucket. Unanimous ($3{:}0$) questions resolve in markedly fewer tokens than disagreement ($1{:}1{:}1$) questions ($2.53\times$ ratio, Spearman $\rho = -0.27$).}

\betterparagraph{Query Sequencing Stage Speedup}
During the query sequencing stage, releasing each answer as soon as its gate is decided lowers the mean TTA from 1507 to 658\,s at the same makespan, and running short sessions first lowers it further to 490\,s.
SPT answers the decided questions $11\%$ earlier on average ($12\%$ at the 90th percentile) and makes the undecided questions ready $8\%$ earlier.
Terminating the remaining calls of decided questions reduces the expert tokens by $9\%$ and ends the expert phase $8\%$ earlier.
Overall, the makespan decreases by $18\%$, and the mean and median TTA decrease by $3.7\times$ and $6.4\times$.

% \begin{figure}[t]
% \centering
% \includegraphics[width=\linewidth]{Figures/MOSAIC_overall_speedup.png}
% \caption{\RK{Can you please share the original figure with and without labels? It is not the omnigraffle file.}\SM{quality of this figure to be improved}End-to-end wall-time decomposition. For each benchmark a
% pair of stacked bars compares the round-robin baseline (left) with
% MOSAIC (right)}
% \label{fig:overall-speedup}
% \end{figure}

\begin{table}[t]
\centering
\small
\caption{End-to-end makespan and mean TTA (seconds and speedup over RR) and accuracy (\%).
The makespan includes expert and aggregation stage. MOSAIC incorporates Adaptive Aggregation and Query Sequening.
}
\label{tab:e2e}
\resizebox{\columnwidth}{!}{%
\begin{tabular}{lrrrrr}
\toprule
 & \multicolumn{2}{c}{Makespan} & \multicolumn{2}{c}{Mean TTA} & \\
\cmidrule(lr){2-3}\cmidrule(lr){4-5}
System & s & Speedup & s & Speedup & Acc. \\
\midrule
\multicolumn{6}{l}{\emph{MMLU-Pro}} \\
RR & 3946 & 1.00$\times$ & 3195 & 1.00$\times$ & 63.52 \\
Expert placement only & 1712 & 2.31$\times$ & 1507 & 2.12$\times$ & 63.05 \\
MOSAIC & 1401 & 2.82$\times$ & 408 & 7.83$\times$ & 62.57 \\
\midrule
\multicolumn{6}{l}{\emph{MedMCQA}} \\
RR & 1341 & 1.00$\times$ & 1146 & 1.00$\times$ & 57.38 \\
Expert placement only & 691 & 1.94$\times$ & 638 & 1.80$\times$ & 57.69 \\
MOSAIC & 490 & 2.74$\times$ & 291 & 3.94$\times$ & 57.16 \\
\midrule
\multicolumn{6}{l}{\emph{GPQA}} \\
RR & 2801 & 1.00$\times$ & 2349 & 1.00$\times$ & 55.05 \\
Expert placement only & 1165 & 2.40$\times$ & 1088 & 2.16$\times$ & 54.63 \\
MOSAIC & 835 & 3.35$\times$ & 484 & 4.85$\times$ & 54.34 \\
\bottomrule
\vspace{-8mm}
\end{tabular}%
}
\end{table}

\betterparagraph{Overall Speedup}
% End-to-end pipeline runtime is the sum of expert-generation and aggregator wall time. MOSAIC
% accelerates the two stages by orthogonal mechanisms: workload-aware
% ILP scheduling on the expert side, and confidence-based gating on the
% aggregator side. Figure~\ref{fig:overall-speedup} reports
% the absolute wall-time decomposition for both schedules on the three
% benchmarks. Combining the two stages, MOSAIC reduces the end-to-end wall time from
% $4149.9$ s to $2427.6$ s on MMLU\_Pro ($\mathbf{1.71\times}$), from
% $1321.0$ s to $648.6$ s on MedMCQA ($\mathbf{2.04\times}$), and from
% $2696.4$ s to $1153.1$ s on GPQA ($\mathbf{2.34\times}$). 
% On MedMCQA, the medical-MCQ questions reach a clear majority among the three
% expert votes most of the time, so the aggregator gate alone gives
% $4.23\times$ and accounts for most of the end-to-end gain. On GPQA
% the R1-dominated workload is where the scheduler pays off most
% ($2.54\times$ expert).
Compared with round-robin (\autoref{tab:e2e}), MOSAIC shortens the makespan by $2.7$--$3.4\times$ and the mean TTA by $3.9$--$7.8\times$.
Expert placement alone shortens the makespan by $1.9$--$2.4\times$ but the mean TTA only by $1.8$--$2.2\times$, because its answers wait for the expert phase (mean TTA at $88$--$93\%$ of the makespan). The query sequencing and adaptive aggregation reduce the makespan by a further $18$--$29\%$ and the mean TTA by $2.2$--$3.7\times$.
Accuracy stays within 1\,pp of RR on MMLU-Pro and MedMCQA, and GPQA, within the benchmark's run-to-run variation.

\section{Conclusions}
\label{sec:conclusion}

% MoA-style inference creates a distinct scheduling problem.
% %
% MOSAIC addresses this setting by jointly optimizing the expert-generation phase and the aggregation phase.
% %
% For expert generation, it uses offline profiling and ILP-based scheduling to balance model load cost, generation cost, and selective replication across GPUs.
% %
% For aggregation, MOSAIC exploits inter-expert agreement as a confidence signal to skip unnecessary aggregator calls. 
% %
% Across benchmarks, it reduces expert-stage makespan and substantially lowers aggregation overhead while preserving accuracy. 
% %
% Overall, MOSAIC demonstrates the need for workload-aware scheduling in MoA serving.

MOSAIC addresses the scheduling problem created by skewed expert demand and variable generation lengths in MoA inference.
It places and replicates experts across GPUs using an ILP, orders each GPU's queries by predicted output lengths, and skips the aggregator when experts agree.
On a 4-GPU server, MOSAIC achieves $2.7$--$3.4\times$ end-to-end speedups and $3.9$--$7.8\times$ lower mean TTA over round-robin scheduling, with accuracy within 1pp.
% Overall, MOSAIC demonstrates the need for workload-aware scheduling in MoA serving.

% MOSAIC addresses the scheduling challenge posed by uneven expert demand and variable generation times in MoA inference. It distributes and replicates experts across GPUs using an ILP, prioritizes work based on predicted output lengths, and skips the aggregator when experts concur. On a 4-GPU server, MOSAIC achieves $2.7$—$3.4\times$ speedups and $3.9$—$7.8\times$ lower mean time-to-answer compared to round-robin scheduling, maintaining accuracy within 1pp. MOSAIC highlights the necessity of workload-aware scheduling in MoA serving.

\section*{Limitations}
% While in MOSAIC, we demonstrate the benefits in accelerating the mixture of model serving, this work faces limitations in the static selection of the `confidence criterion'. In specific, for any task type, the user needs to select the aggregation skip criteria in advance that does not change during the inference serving stage. Static `confidence selection' along with associated hyperparameter choices of thresholding remains as a limitation of the current work. Further, the ILP solver can take higher benefit for tasks that has categorical partitioning based on task sub-type or criticality (example: MMLU-Pro), compared to the tasks that has no further sub-partitioning (example: GPQA).

% \SM{Alternative to the first paragraph above: 
MOSAIC targets offline, batch-oriented workloads in which queries and their routed experts are known before execution; it re-solves cheaply per burst (Appendix~\ref{appendix:dynamic}) but does not handle per-request latency SLOs, preemption, or admission control as in online serving.
Our hardware experiments use a single 4-GPU node with six 7--9B experts; larger fleets are evaluated only with a calibrated synthetic study (\autoref{appendix:scaling}).
% The aggregation gate uses a static, task-dependent confidence criterion and threshold chosen in advance; plurality voting applies only to tasks with a discrete answer set, and the entropy-based extension to open-ended outputs is analysed but not yet evaluated end to end \SM{drop this clause if E4 is done}.
% Finally, the ILP benefits more from tasks with categorical sub-partitions (e.g., MMLU-Pro) than from tasks without them (e.g., GPQA).
Query sequencing ranks output lengths from each expert's hidden states, which requires white-box access to the experts and a per-expert predictor fitted on calibration data; so length-aware query sequencing does not apply as-is to API-served models.
% }

% Our MoA serving acceleration may benefit the expert model based task processing in an Agentic framework. However,  extending this work of LLMs-based monolithic mixture-of-agent serving may not remain straightforward to heterogeneous agentic workload execution as many of the tasks may not be bottle-necked by the LLM inference execution latency. Further investigation on fruitful adoption of MOSAIC in such workloads remains as an interesting future research. 

\section*{LLM Usage}
Generative AI tools (Gemini, Claude) were used to assist in writing scripts for plotting results and automating the experimental workflow. All core research code was developed solely by the authors.

\bibliography{ref} % Points to your ref.bib file

\appendix

\section{Appendix}

% \insertFigure{dp}{Overview figure.}
\subsection{The Pitfalls of Data Concurrency and Over-Sharding}
\label{appendix:dp_shard}
% \SK{This is too detailed ... you need to make it precise and convert to 1/2 of current size.}
In data parallelism (DP), a complete copy of each expert is loaded on all 4 GPUs. Prompts requiring that expert are evenly shared across workers. Experts are processed sequentially with a sync barrier between them.
This strategy differs from expert-per-worker or round-robin placement in critical path: runs each worker's assigned experts serially with model loads overlapped across workers, giving $\text{expert\_latency} \approx \max_{w}(\sum_{m} \text{load} + \text{gen})$ and exposing within-worker stragglers as idle time.
DP serializes over models but parallelizes within each, giving $\text{expert\_latency} \approx \sum_{m} \max_{w}(\text{load} + \text{gen})$ and replacing straggler-idle with a replicated load tax (every model loaded on every worker).
\autoref{fig:popularity2-shrinked}a confirms the tradeoff; DP cuts the \textasciitilde45\% idle share of roundrobin runtime to give a overall \textasciitilde9\% wall-time improvement in default model-pool, and similarly improves allR (7080s to 6431s).
In allnonR, however, DP is \textasciitilde29\% slower (631s vs 892s), the per-expert generate budget is too small to amortize serialized loads.

DP incurs a second cost due to throughput collapse from over-sharding.
vLLM\citep{kwon2023efficient} relies on continuous batching, which only reaches peak tok/s when enough prompts are in flight to keep the model fully utilized.
Below a model and GPU-dependent threshold, throughput scales sub-linearly with batch size, and at very small batches, the GPU starves.
\autoref{fig:popularity2-shrinked}b makes this visible by plotting the throughput of each model in the default model-pool for both roundrobin (full data shard) and DP (1/4 shard).
The full-shard points sit at the right for each color (expert), showing peak-throughput.
The DP points are located on the left (137-239 prompts per shards) and throughput is reduced by 40-60\% for Qwen, Exaone, and GLM.
Gemma is an exception because skill-weighted recruitment over-picked it in the default pool, even its 1/4 shard (\textasciitilde408 prompts) stays above its saturation threshold. Thus, DP barely costs throughput for Gemma.
Reasoning models cluster around 1000-2000 tok/s, regardless of shard size.
Their long autoregressive decode phase is memory-bandwidth bound, so a handful of in-flight prompts already saturate HBM and additional batch yields little.

Taken together, DP trades off straggler-idle time for two new costs: serialized load and reduced throughput per shard.
It only wins when the per-expert generate budget is large enough to absorb both.
Expert-per-worker placement, by contrast, operates at peak per-shard throughput and pays the model load only once per worker.
Its sole weakness is idle time from popularity and cost-skew.
%
% Next, we develop a scheduler that assigns cost-aware solvers to workers, balancing the load and closing idle gaps.

\subsection{Full Model Pool}
Table~\ref{tab:full_model_pool} lists the full set of models considered in our study, separated into reasoning and non-reasoning models.

\begin{table*}[t]
\centering
\small
\setlength{\tabcolsep}{6pt}
\renewcommand{\arraystretch}{0.95}
\begin{tabular}{@{}l c l@{}}
\toprule
\textbf{Model Name} & \textbf{Size} & \textbf{Hugging Face Checkpoint} \\
\midrule
\multicolumn{3}{@{}l}{\textbf{Non-reasoning models}} \\
\midrule
BioLlama \citep{contactdoctor_biomedical_llama3_8b} & 8B & \href{https://huggingface.co/ContactDoctor/Bio-Medical-Llama-3-8B}{ContactDoctor/Bio-Medical-Llama-3-8B}
\\
DeepSeekMath \citep{shao2024deepseekmath} & 7B & \href{https://huggingface.co/deepseek-ai/deepseek-math-7b-instruct}{deepseek-ai/deepseek-math-7b-instruct}
\\
Exaone \citep{exaone-3.5} & 7.8B & \href{https://huggingface.co/LGAI-EXAONE/EXAONE-3.5-7.8B-Instruct}{LGAI-EXAONE/EXAONE-3.5-7.8B-Instruct} 
\\
Gemma2 \citep{gemma_2024} & 9B & \href{https://huggingface.co/google/gemma-2-9b-it}{google/gemma-2-9b-it}
\\
GLM4 \citep{glm2024chatglm} & 9B & \href{https://huggingface.co/THUDM/glm-4-9b}{THUDM/glm-4-9b} \\
Granite \citep{granite31_8b_instruct_hf} & 8B & \href{https://huggingface.co/ibm-granite/granite-3.1-8b-instruct}{ibm-granite/granite-3.1-8b-instruct}
\\
InternLM3 \citep{cai2024internlm2} & 8B & \href{https://huggingface.co/internlm/internlm3-8b-instruct}{internlm/internlm3-8b-instruct}
\\
Llama3.1 \citep{grattafiori2024llama} & 8B & \href{https://huggingface.co/meta-llama/Llama-3.1-8B-Instruct}{meta-llama/Llama-3.1-8B-Instruct}
\\
Mathstral \citep{mathstral_7b_v01_hf} & 7B & \href{https://huggingface.co/mistralai/Mathstral-7B-v0.1}{mistralai/Mathstral-7B-v0.1}
\\
Mistral \citep{mistralnemo_instruct_2407_hf} & 12B & \href{https://huggingface.co/mistralai/Mistral-Nemo-Instruct-2407}{mistralai/Mistral-Nemo-Instruct-2407}
\\
Phi3.5-mini \citep{abdin2024phi3technicalreporthighly} & 3.5B & \href{https://huggingface.co/microsoft/Phi-3.5-mini-instruct}{microsoft/Phi-3.5-mini-instruct}
\\
Qwen2.5 \citep{qwen2.5} & 7B & \href{https://huggingface.co/Qwen/Qwen2.5-7B-Instruct}{Qwen/Qwen2.5-7B-Instruct}
\\
Qwen2.5-Coder \citep{hui2024qwen2} & 7B & \href{https://huggingface.co/Qwen/Qwen2.5-Coder-7B-Instruct}{Qwen/Qwen2.5-Coder-7B-Instruct}
\\
Qwen2.5-Math \citep{yang2024qwen25mathtechnicalreportmathematical} & 7B & \href{https://huggingface.co/Qwen/Qwen2.5-Math-7B-Instruct}{Qwen/Qwen2.5-Math-7B-Instruct}
\\
\midrule
\multicolumn{3}{@{}l}{\textbf{Reasoning models}} \\
\midrule
LlamaR1 \citep{deepseekai2025deepseekr1incentivizingreasoningcapability} & 8B & \href{https://huggingface.co/deepseek-ai/DeepSeek-R1-Distill-Llama-8B}{deepseek-ai/DeepSeek-R1-Distill-Llama-8B}
\\
QwenR1 \citep{deepseekai2025deepseekr1incentivizingreasoningcapability} & 7B & \href{https://huggingface.co/deepseek-ai/DeepSeek-R1-Distill-Qwen-7B}{deepseek-ai/DeepSeek-R1-Distill-Qwen-7B}
\\
DLERR1 \citep{liu2025dler} & 7B & \href{https://huggingface.co/nvidia/DLER-R1-7B-Research}{nvidia/DLER-R1-7B-Research}
\\
ThinkerR1 \citep{guha2025openthoughtsdatarecipesreasoning} & 7B & \href{https://huggingface.co/open-thoughts/OpenThinker-7B}{open-thoughts/OpenThinker-7B}
\\
StratosR1 \citep{bespoke_stratos_7b_hf} & 7B & \href{https://huggingface.co/bespokelabs/Bespoke-Stratos-7B}{bespokelabs/Bespoke-Stratos-7B}
\\
SkyworkR1 \citep{he2025skywork} & 7B & \href{https://huggingface.co/Skywork/Skywork-OR1-7B}{Skywork/Skywork-OR1-7B}
\\
\bottomrule
\end{tabular}
\caption{The model pool considered for MMLU-Pro, MedMCQA, GPQA for Skill-MoE type MoA workload.}
\label{tab:full_model_pool}
\end{table*}

\subsection{Hybrid Model Pool Used in Evaluation}
\label{appendix:pool}

The six experts that constitute the hybrid pool used throughout
Section~\ref{sec:results} are listed in
Table~\ref{tab:appendix:pool}. Two of the six are reasoning-tuned and
emit long $\langle\text{think}\rangle$ rationales; the remaining four
are instruction-tuned and produce concise answers. The mean output
length differs by roughly an order of magnitude across the two
families, which is the heterogeneity our scheduler is designed to
absorb.

Each expert runs as a single vLLM session at tensor-parallel rank 1.
% Per-session load cost $\ell_m$ ranges between 38--47\,s and is dominated by weight transfer and KV-cache pre-allocation on a single A100 80\,GB.
Measured cold-start load costs $\ell_m$ are 35.2\,s (LlamaR1), 36.2\,s (Qwen), 37.3\,s (Exaone), 37.9\,s (QwenR1), 40.4\,s (Gemma) and 41.7\,s (GLM), rising to 50--64\,s when several workers load concurrently; they are dominated by weight transfer and KV-cache pre-allocation on a single A100 80\,GB.

\begin{table}[h]
\centering
\caption{Six-expert hybrid pool. Output statistics are measured on
the MMLU\_Pro test set and rounded to representative values; analogous
ranges hold on MedMCQA.}
\label{tab:appendix:pool}
\small
\begin{tabular}{llr}
\toprule
Expert & Family & Mean output (tok) \\
\midrule
LlamaR1 & reasoning-tuned & $\sim 3\,000$ \\
QwenR1  & reasoning-tuned & $\sim 3\,000$ \\
\midrule
Gemma   & instruction-tuned & $\sim 200$ \\
Exaone  & instruction-tuned & $\sim 200$ \\
GLM     & instruction-tuned & $\sim 200$ \\
Qwen    & instruction-tuned & $\sim 200$ \\
\bottomrule
\end{tabular}
\end{table}

\subsection{Datasets}
\label{appendix:datasets}

All three benchmarks are used in their standard test splits. Each
test question is routed to three experts by the upstream skill-based
recruit, producing a per-expert slot count $N_m$ summarised in
Table~\ref{tab:appendix:datasets}.

\begin{table}[h]
\centering
\caption{Per-expert prompt slots on the three benchmarks at seed 0.
The three workloads exercise very different recruit shapes:
MMLU\_Pro spreads slots across all six experts, MedMCQA concentrates
on the instruction-tuned family, and GPQA is dominated by LlamaR1.}
\label{tab:appendix:datasets}
\small
\resizebox{\columnwidth}{!}{
\begin{tabular}{lrrr}
\toprule
Expert  & MMLU\_Pro $N_m$ & MedMCQA $N_m$ & GPQA $N_m$ \\
\midrule
LlamaR1 & 1\,070 &     620 & 429 \\
QwenR1  & 1\,428 &     144 & 158 \\
Gemma   & 1\,632 & 4\,195 &   4 \\
Exaone  &    547 &     625 &   3 \\
GLM     &    667 & 1\,283 &   0 \\
Qwen    &    956 & 5\,682 &   0 \\
\midrule
\textbf{Total prompts} & \textbf{6\,300} & \textbf{12\,549} & \textbf{594} \\
\textbf{Test questions} & 2\,100 & 4\,183 & 198 \\
\bottomrule
\end{tabular}
}
\end{table}

MMLU\_Pro contains 2\,100 multiple-choice questions across 14 STEM
and humanities categories. MedMCQA contributes 4\,183 medical
multiple-choice questions. GPQA contributes 198 graduate-level
science questions whose narrow skill profile drives the recruiter to
concentrate $\sim$$72\%$ of its slots on LlamaR1; the two
lowest-scoring experts (GLM, Qwen) are never selected. All three
benchmarks are served by the same six-expert hybrid pool described
in Appendix~\ref{appendix:pool}.
The instance-level skill-based router of \citet{chen2025symbolic} is used unmodified as an offline upstream stage: for each question it scores experts against symbolic skill descriptions and recruits the top $k{=}3$. Its only interface to MOSAIC is the per-expert prompt count $N_m$. We will release the question-to-expert assignments, calibration statistics and scheduler code.

\subsection{Multi-seed Validation}
\label{appendix:multiseed}

Each configuration is repeated over three random seeds (0, 1, 2) to
control for the stochastic component of vLLM's continuous-batching
scheduler. Table~\ref{tab:appendix:multiseed} reports the per-seed
expert-phase wall time of MOSAIC on the three benchmarks.

\begin{table}[h]
\centering
\caption{Per-seed wall time of MOSAIC. All seed wall times are within $\pm 6\%$ of the seed-0 value, confirming that the
schedule is not over-fit to a particular recruit realisation. }
\label{tab:appendix:multiseed}
\small
\begin{tabular}{lrrr}
\toprule
Dataset & Seed 0 (s) & Seed 1 (s) & Seed 2 (s) \\
\midrule
MMLU\_Pro & 1601  & 1668 & 1637 \\
MedMCQA   &  585  &  551  &  527  \\
GPQA      &  925  & 945 & 896 \\
\bottomrule
\end{tabular}
\end{table}

\subsection{Choice of the Per-Worker Cap $K$}
\label{appendix:kcap}
% <paste the moved paragraph here>
The capacity cap $K{=}2$ adopted throughout is the smallest value
that satisfies the two requirements implied by
\autoref{eq:c:rho-replica-cap}. First, a feasibility lower bound
$K \ge \lceil M/N \rceil = 2$ holds independently of any calibration.
Second, the per-model replica caps measured from calibration sit at
$R_m^{\max}{=}N$ for the two reasoning experts ($\rho_m \approx
0.02$, since their generation time dwarfs the load cost) but
$R_m^{\max}{=}1$ for the four instruction-tuned experts ($\rho_m \in
[0.2, 0.95]$, where an extra load would consume most of the savings).
$K{=}2$ is therefore precisely the capacity that grants the reasoning
experts the room they need to split without admitting a second load
of any instruction-tuned expert: $K{=}1$ is infeasible and $K{=}3$
would over-replicate models whose $R_m^{\max}{=}1$ already forbids
the extra load. The solver treats $K{=}2$ as a soft budget,
saturating it on workers that benefit from a reasoning slice and
leaving it slack elsewhere; this is most pronounced on GPQA, where
the recruiter sends $72\%$ of all prompts to LlamaR1 and MOSAIC
consequently replicates LlamaR1 across three workers, yielding
the largest observed speedup, because RR's single-worker
LlamaR1 allocation is the bottleneck.

\subsection{Length Predictor and Its Generalization}
\label{appendix:predictor}
For each expert, the predictor is a ridge regression (with the regularization strength chosen by cross-validation) from the hidden state of the last prompt token at $3/4$ depth to the log output length, trained on the expert's calibration run.
On MMLU-Pro, its held-out Spearman correlation is $0.72$ (QwenR1), $0.75$ (LlamaR1), $0.68$ (Gemma), $0.72$ (Qwen), $0.69$ (GLM), and $0.73$. Repeated sampling of the same prompts bounds any prompt-only predictor at $0.86$--$0.89$.
Features matter more than the regressor: for QwenR1, the question category reaches $0.37$, a sentence embedding of the prompt $0.58$, and entropy statistics of the first 128 generated tokens $0.22$--$0.39$.

\begin{figure}[t]
    \centering
    \includegraphics[width=\linewidth]{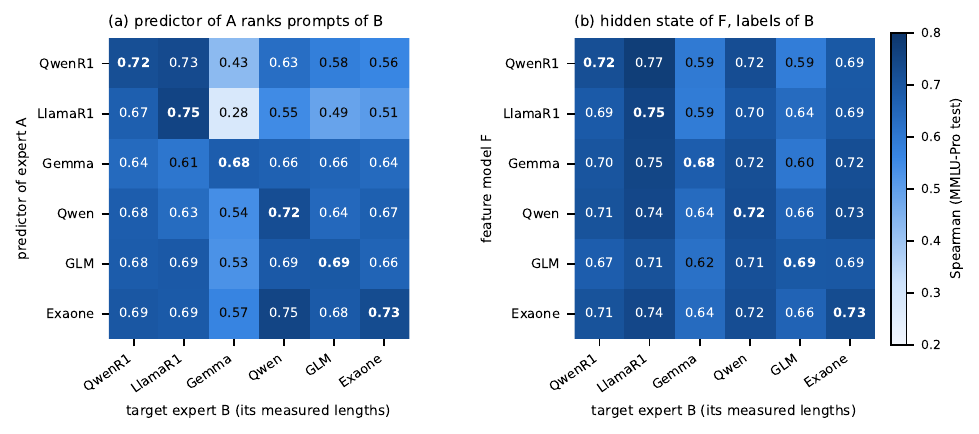}
    \vspace{-7mm}
    \caption{Cross-expert generalization on MMLU-Pro (Spearman correlation). (a) The predictor of expert A ranks the prompts of expert B. (b) The hidden state of model F combined with the length labels of expert B; the diagonal uses each expert's own hidden state.}
    \vspace{-4mm}
    \label{fig:fig_length_prediction_generalization}
\end{figure}

\paragraph{The generalization of length predictor}
The predictor is trained per expert and per task.
\autoref{tab:gen} trains it on the calibration set of one task and tests it on another.
Transfer across tasks loses $0.05$--$0.25$ in Spearman correlation, but a single predictor trained on the pooled calibration sets of all three tasks is as accurate as the in-task predictor (within $\pm0.04$), so the predictor can be calibrated once per expert.
Rankings also transfer across experts (\autoref{fig:fig_length_prediction_generalization}): the predictor of one reasoning expert ranks the prompts of the other with a correlation of $0.67$--$0.73$, and the hidden state of a single model, combined with each expert's own length labels, predicts every expert about as well as the expert's own hidden state ($0.59$--$0.77$), so features need to be extracted only once per prompt.
A Spearman correlation above $0.6$ suffices to obtain $62$--$91\%$ of the within-session SPT gain achievable with the true lengths.

\begin{table}[h]
\centering
\small
\caption{Spearman correlation between predicted and measured output lengths on each test set, by calibration source. Pooled: all three calibration sets combined.}
\label{tab:gen}
\resizebox{\columnwidth}{!}{%
\begin{tabular}{llrrrr}
\toprule
Expert & Test set & MMLU-Pro & MedMCQA & GPQA & Pooled \\
\midrule
QwenR1 & MMLU-Pro & 0.72 & 0.47 & 0.68 & \textbf{0.74} \\
LlamaR1 & MMLU-Pro & \textbf{0.75} & 0.70 & 0.67 & \textbf{0.75} \\
LlamaR1 & GPQA & 0.60 & 0.56 & 0.66 & \textbf{0.67} \\
Gemma & MMLU-Pro & \textbf{0.68} & 0.58 & 0.46 & \textbf{0.68} \\
Gemma & MedMCQA & 0.69 & 0.71 & 0.62 & \textbf{0.72} \\
Qwen & MMLU-Pro & 0.72 & 0.61 & 0.47 & \textbf{0.73} \\
Qwen & MedMCQA & 0.47 & \textbf{0.59} & 0.38 & \textbf{0.59} \\
GLM & MedMCQA & 0.45 & \textbf{0.56} & 0.29 & 0.52 \\
\bottomrule
\end{tabular}%
}
\end{table}
% \insertFigure{fig_length_prediction_generalization}{Cross-expert generalization on MMLU-Pro (Spearman correlation). (a) The predictor of expert A ranks the prompts of expert B. (b) The hidden state of model F combined with the length labels of expert B; the diagonal uses each expert's own hidden state.}

\subsection{Session Order and Question Completion}
\label{appendix:session_order}
\autoref{tab:session-order} reports the measurements behind Observation~4 (\autoref{sec:motivation}) and the session order of the query-sequencing stage (\autoref{sec:qsched}).
Relative to the placement's own order, running the short instruction-tuned sessions first reduces the median time until all three expert answers of a question are available from 517 to 335\,s, and the median time until two of them agree, which decides the question under the aggressive gate, from 260 to 217\,s; running the reasoning sessions first delays both beyond 1300\,s.
\begin{table}[h]
\centering
\small
\caption{Completion of MMLU-Pro questions (measured run with a fixed expert placement; seconds since the start of the batch). The rows differ only in the order of the sessions on each GPU; since every session is an independent engine, each expert call takes the same time and the makespan is identical.}
\label{tab:session-order}
\resizebox{\columnwidth}{!}{%
\begin{tabular}{lrrr}
\toprule
Session order on each GPU & 3 answers & 2 answers & Makespan \\
 & (median) & agree (median) & \\
\midrule
Reasoning sessions first & 1376 & 1330 & 1490 \\
Placement's own order & 517 & 260 & 1490 \\
Short sessions first & \textbf{335} & \textbf{217} & 1490 \\
\bottomrule
\end{tabular}%
}
\end{table}

\subsection{ILP vs.\ Cost-Aware Greedy Scheduling}
\label{appendix:greedy}
% paste Table 2 here unchanged (keep \label{tab:greedy}); change [t] to [h] or [H]
\begin{table}[h]
\centering
\small
\caption{
% \SM{additional table. MedMCQA greedy cells need E2;
% % all cells  become mean +- CI after E1.
% If the length-prediction ILP is adopted, add
% a column "MOSAIC+LP" AND rerun Greedy with the same predictor (E3).}
Expert-stage wall time (s) and speedup over RR. LPT uses the same calibration as MOSAIC.
% ``cat.'' schedules whole (expert, category) blocks instead of divisible per-expert pools.
}
\label{tab:greedy}
\resizebox{\columnwidth}{!}{%
\begin{tabular}{lcrrr}
\toprule
Task & GPUs & RR & LPT greedy & MOSAIC (ILP) \\
\midrule
MMLU-Pro        & 4 & 2635 & 1678 (1.57$\times$) & \textbf{1557 (1.69$\times$)} \\
MMLU-Pro        & 3 & 2779 & 2577 (1.08$\times$) & \textbf{2409 (1.15$\times$)} \\
% MMLU-Pro (cat.) & 4 & 2635 & 2047 (1.29$\times$) & \textbf{1539 (1.71$\times$)} \\
% MedMCQA         & 4 & \SM{805} & \SM{E2} & \textbf{\SM{527} (1.53$\times$)} \\
GPQA            & 4 & 2348 &  946 (2.48$\times$) & \textbf{935 (2.51$\times$)} \\
GPQA            & 3 & 2340 & 1768 (1.32$\times$) & \textbf{1330 (1.76$\times$)} \\
\bottomrule
\end{tabular}%
}
\end{table}

LPT uses the same calibrated costs but places work greedily, so RR$\rightarrow$LPT isolates the benefit of cost-awareness while LPT$\rightarrow$ILP isolates the advantage of search (\autoref{tab:greedy}).
% ... rest of the paragraph unchanged (L238–242), with these edits:
%   - drop the \paragraph{ILP vs.\ cost-aware greedy.\SM{move to appendix}} heading
%   - drop \label{sec:eval:greedy}  (the subsection has its own label)
%   - "(\autoref{appendix:scaling})" -> "(Appendix~\ref{appendix:scaling})"
With 4 GPUs, cost-awareness yields most of the gain (LPT $1.57\times$/$2.48\times$ vs.\ ILP $1.69\times$/$2.51\times$ on MMLU-Pro/GPQA), since splitting each reasoning expert across two GPUs is near-optimal and greedy already finds it.
% The picture changes when GPUs are scarce.
Under tighter GPU constraints, however, greedy heuristics break down.
With 3 GPUs, each extra copy costs a full model load, and LPT's early, irrevocable placement of the heaviest expert falls short: the ILP reaches $1.76\times$ vs.\ $1.32\times$ on GPQA and $1.15\times$ vs.\ $1.08\times$ on MMLU-Pro.
In synthetic scaling up to 80 experts and 64 workers (Appendix~\ref{appendix:scaling}), the ILP consistently outperforms the strongest greedy baseline, reducing makespan by $7.7\%$, $13.6\%$ and $17\%$ at $M{=}6, 20, 40$, while CP-SAT stays within $6.1\%$ of the optimality bound after $30$\,s.
Thus, while cost-awareness provides the baseline speedup on moderately provisioned clusters, exact ILP optimization becomes critical under resource scarcity or at larger cluster scales.
In short, cost-awareness yields most of the gain on moderately provisioned clusters, while the ILP becomes critical under resource scarcity or at larger fleet size.

\subsection{ILP Scalability and Solve Time}
\label{appendix:scaling}
We sweep the number of experts $M$ and workers $W$ synthetically. Per-expert parameters are resampled, with lognormal jitter, from the fitted throughput models of our eight profiled experts ($R^2 \ge 0.995$), and the workload is scaled to the paper's per-GPU pressure. A regression test pins the generator to the exact CP-SAT program used on hardware (534 solves; the real MMLU-Pro instance solves in $0.03$\,s).
\autoref{tab:scaling} reports the optimality gap after a $30$\,s solve budget. Across all instances, the best greedy baseline's makespan is higher than the ILP's (median $+22.7\%$ 
% \SM{check: per-M values are $\le 17\%$}
).

\begin{table}[h]
\centering
\small
\caption{CP-SAT optimality gap after 30\,s (0 = proven optimal; time to optimality in parentheses).}
\label{tab:scaling}
\resizebox{\columnwidth}{!}{%
\begin{tabular}{lrrrrr}
\toprule
 & $W{=}4$ & $W{=}8$ & $W{=}16$ & $W{=}32$ & $W{=}64$ \\
\midrule
$M{=}6$  & 0 (0.09\,s) & 0     & 0.7\% & 0.3\% & 0.2\% \\
$M{=}10$ & 0 (0.33\,s) & 0.7\% & 1.2\% & 0.7\% & 0.4\% \\
$M{=}20$ & 0 (6.4\,s)  & 0.5\% & 1.4\% & 1.6\% & 1.1\% \\
$M{=}40$ & 0           & 0.2\% & 1.5\% & 2.3\% & 2.8\% \\
$M{=}80$ & 0           & 0     & 1.3\% & 3.1\% & 6.1\% \\
\bottomrule
\end{tabular}%
}
\end{table}

\begin{table}[h]
\centering
\small
\caption{Per-burst CP-SAT solve time (ms, mean of 5 runs) by pool size $M$ (cap $K$) and burst size $B$.}
\label{tab:burst-solve}
\begin{tabular}{lrrrr}
\toprule
$M$ ($K$) & $B{=}50$ & $200$ & $800$ & $2100$ \\
\midrule
6 (2)  &  6.8 & 10.5 & 19.0 &  27.3 \\
8 (2)  & 11.7 &  9.2 &  8.9 &   9.0 \\
10 (3) & 11.1 & 76.0 & 74.6 & 156.9 \\
12 (3) & 18.9 & 82.0 & 48.3 &  51.4 \\
\bottomrule
\end{tabular}
\end{table}

\subsection{Burst-wise and Mixed Workloads}
\label{appendix:dynamic}
MOSAIC does not need the whole workload up front: it can re-solve the ILP for each burst of requests as it arrives, which takes at most $27$\,ms at our scale (\autoref{tab:burst-solve}), below $0.004\%$ of the entire generation time.
\autoref{tab:dynamic} evaluates this mode.
The left panel dispatches MMLU-Pro in bursts of $B$ questions and re-solves per burst; the expert-stage speedup over RR grows with burst size, from $1.22\times$ ($B{=}400$) to $1.36\times$ ($B{=}800$) and $1.69\times$ for the full batch.
The right panel mixes MMLU-Pro and GPQA at GPQA fraction $\rho$ in bursts of $400$ and compares per-burst re-solving with a static placement fixed at initialization.
% \SM{define: RR or ILP, and for which mix?}.
Re-solving is $1.32\times$ ($\rho{=}0$) to $3.13\times$ ($\rho{=}1$) faster, because a static placement leaves GPUs idle as demand shifts toward the reasoning experts.

% In \autoref{tab:dynamic}, the left panel evaluates MMLU-Pro dispatched in bursts of $B$ questions and re-solves the ILP for each burst. The right panel mixes MMLU-Pro and GPQA at GPQA fraction $\rho$ in bursts of $400$, comparing dynamic per-burst re-solving against a static placement at initialization.
% % \SM{define: fixed for which mix? RR or ILP?}.

\begin{table}[h]
\centering
\small
\caption{Expert-stage wall time (s). Left: burst size $B$ vs.\ RR on MMLU-Pro. Right: mixed MMLU-Pro/GPQA bursts vs.\ a fixed placement.}
\label{tab:dynamic}
\resizebox{\columnwidth}{!}{%
\begin{tabular}{rrrr|rrrr}
\toprule
$B$ & MOSAIC & RR & Speedup & $\rho$ & MOSAIC & Fixed & Speedup \\
\midrule
400  & 742.6  & 906.9  & 1.22$\times$ & 0.00 &  679.5 &  894.9 & 1.32$\times$ \\
800  & 886.3  & 1208.6 & 1.36$\times$ & 0.25 & 1011.4 & 1822.6 & 1.80$\times$ \\
2100 & 1651.4 & 2648.3 & 1.60$\times$ & 0.50 & 1436.5 & 2658.5 & 1.85$\times$ \\
     &        &        &              & 0.75 & 1461.3 & 3585.2 & 2.45$\times$ \\
     &        &        &              & 1.00 & 1523.4 & 4763.1 & 3.13$\times$ \\
\bottomrule
\end{tabular}%
}
\end{table}

% \begin{table}[h]
% \centering
% \small
% \caption{Expert-stage wall time (s). Left: burst size $B$ vs.\ RR on MMLU-Pro. Right: mixed MMLU-Pro/GPQA bursts vs.\ a fixed placement.}
% \label{tab:dynamic}
% \resizebox{\columnwidth}{!}{%
% \begin{tabular}{rrrr@{\hspace{10pt}}rrrr}
% \toprule
% $B$ & MOSAIC & RR & Speedup & $\rho$ & MOSAIC & Fixed & Speedup \\
% \midrule
% 400  & 742.6  & 906.9  & 1.22$\times$ & 0.00 &  679.5 &  894.9 & 1.32$\times$ \\
% 800  & 886.3  & 1208.6 & 1.36$\times$ & 0.25 & 1011.4 & 1822.6 & 1.80$\times$ \\
% 2100 & 1651.4 & 2648.3 & 1.60$\times$ & 0.50 & 1436.5 & 2658.5 & 1.85$\times$ \\
%      &        &        &              & 0.75 & 1461.3 & 3585.2 & 2.45$\times$ \\
%      &        &        &              & 1.00 & 1523.4 & 4763.1 & 3.13$\times$ \\
% \bottomrule
% \end{tabular}%
% }
% \end{table}

\subsection{Robustness of Offline Calibration}
\label{appendix:robust}
We rebuild the ILP schedule from (i) random calibration subsets of size $N_{\text{cal}}$ (20 seeds), (ii) calibration costs perturbed by mean-preserving lognormal noise of scale $\sigma$ (30 seeds), and (iii) another benchmark's calibration, and report the wall-time change relative to the schedule built from the full, clean, native calibration.

\begin{table}[H]
\centering
\small
\caption{Schedule slowdown (\%) vs.\ the full-calibration schedule. Negative values are faster.}
\label{tab:robust}
\setlength{\tabcolsep}{5pt}
\begin{tabular}{lrrr}
\toprule
Setting & MMLU-Pro & GPQA & MedMCQA \\
\midrule
$N_{\text{cal}}{=}25$  & $+7.7$ & $+0.4$ & $+5.5$ \\
$N_{\text{cal}}{=}50$  & $+1.4$ & $0.0$  & $+6.5$ \\
$N_{\text{cal}}{=}100$ & $-0.5$ & $-0.1$ & $+6.0$ \\
$N_{\text{cal}}{=}175$ & $-0.5$ & $0.0$  & $+4.5$ \\
\midrule
$\sigma{=}5\%$  & $-0.3$  & $-0.3$ & $+4.9$ \\
$\sigma{=}10\%$ & $+2.9$  & $0.0$  & $+5.1$ \\
$\sigma{=}20\%$ & $+6.2$  & $+1.7$ & $+7.8$ \\
$\sigma{=}30\%$ & $+14.1$ & $+2.1$ & $+5.6$ \\
\midrule
\multicolumn{4}{l}{Cross-benchmark (target $\leftarrow$ source):} \\
\multicolumn{4}{l}{\quad GPQA $\leftarrow$ MMLU-Pro: $-2.5$} \\
\multicolumn{4}{l}{\quad GPQA $\leftarrow$ MedMCQA: $+0.4$} \\
\multicolumn{4}{l}{\quad MMLU-Pro $\leftarrow$ MedMCQA: $-1.2$} \\
\bottomrule
\end{tabular}
\end{table}

% \begin{table}[h]
% \centering
% \small
% \caption{Schedule slowdown (\%) vs.\ the full-calibration schedule. Negative values are faster.}
% \label{tab:robust}
% \begin{tabular}{lrrr}
% \toprule
% Setting & MMLU-Pro & GPQA & MedMCQA \\
% \midrule
% $N_{\text{cal}}{=}25$  & $+7.7$ & $+0.4$ & $+5.5$ \\
% $N_{\text{cal}}{=}50$  & $+1.4$ & $0.0$  & $+6.5$ \\
% $N_{\text{cal}}{=}100$ & $-0.5$ & $-0.1$ & $+6.0$ \\
% $N_{\text{cal}}{=}175$ & $-0.5$ & $0.0$  & $+4.5$ \\
% \midrule
% $\sigma{=}5\%$  & $-0.3$  & $-0.3$ & $+4.9$ \\
% $\sigma{=}10\%$ & $+2.9$  & $0.0$  & $+5.1$ \\
% $\sigma{=}20\%$ & $+6.2$  & $+1.7$ & $+7.8$ \\
% $\sigma{=}30\%$ & $+14.1$ & $+2.1$ & $+5.6$ \\
% \midrule
% \multicolumn{4}{l}{Cross-benchmark: GPQA$\leftarrow$MMLU-Pro $-2.5$; GPQA$\leftarrow$MedMCQA $+0.4$;} \\
% \multicolumn{4}{l}{\hspace{6.6em}MMLU-Pro$\leftarrow$MedMCQA $-1.2$} \\
% \bottomrule
% \end{tabular}
% \end{table}

The schedule is highly robust to calibration fidelity. Using only $50$ calibration questions reproduces the full-calibration schedule identically on GPQA and comes within $1.4\%$ on MMLU-Pro; $10\%$ multiplicative noise on the per-model cost estimates increases makespan by at most $2.9\%$ on both tasks; and cross-benchmark transfer (scheduling one task from another task's profiles) changes wall time by only $-2.5\%$ to $+0.4\%$.
% \SM{MedMCQA sits at a ~5\% offset in every setting; explain (E8) before submission.}

\subsection{Adaptive Aggregation Algorithm}
\label{appendix:agg_algo}
\begin{algorithm}[H]
\caption{Confidence-Aware Adaptive Aggregation}
\label{alg:adaptive-agg}
\begin{algorithmic}[1]
\Require question $q$, experts $E_1,\dots,E_k$ from skill recruitment, threshold $\tau$
\State $a_i \gets \textsc{Extract}(E_i(q))$ for $i=1,\dots,k$
\State $C \gets \max_{a} \;|\{i : a_i = a\}|\,/\,k$
\If{$C \ge \tau$}
    \State \Return $\operatorname{majority}(\{a_i\})$ \Comment{aggregator is skipped}
\Else
    \State \Return $\operatorname{Aggregator}(q, \{a_i\})$
\EndIf
\end{algorithmic}
\end{algorithm}

\newpage
% ---------------------------------------------------------------------
% BLOCK G5 | Sections/08_Appendix.tex | NEW subsection | appendix only
% Entropy confidence signal (rebuttal R3T4/R3T5). If E4 is done, add a
% third table with Delta-pp / skip% / aggregator wall for the entropy gate.
% ---------------------------------------------------------------------
\subsection{Entropy as a Confidence Signal
% \SM{new app. If E4 is done, add a third table with Delta-pp / skip\% / aggregator wall for the entropy gate.}
}
\label{appendix:entropy}
For each response we compute the entropy of the top-20 next-token distribution at every generated token, average it over the response, and normalize by the expert's global mean. \autoref{tab:entropy} shows that correct responses have lower normalized entropy than incorrect ones, including on AIME24, whose answers are free-form integers. Within the $2{:}1$ bucket, the absolute entropy difference between the two agreeing experts is larger when their shared answer is wrong.

\begin{table}[h]
\centering
\small
\caption{Normalized sequence entropy. Top: correct vs.\ incorrect responses. Bottom: $|\Delta|$ entropy between the two agreeing experts in $2{:}1$ questions for MedMCQA.}
\label{tab:entropy}
\begin{tabular}{lrr}
\toprule
Task & Correct & Incorrect \\
\midrule
AIME24  & 0.787 & 1.091 \\
MedMCQA & 0.954 & 0.991 \\
\midrule
$|\Delta|$ in $2{:}1$ & 0.173 & 0.257 \\
\bottomrule
\end{tabular}
\end{table}

% \bibliography{custom}

\end{document}